\documentclass[letterpaper, 10 pt, conference]{ieeeconf}  % Comment this line out if you need a4paper
\IEEEoverridecommandlockouts       

\usepackage{math}
\usepackage{amsmath}
\usepackage{cite}
\usepackage{hyperref}       % hyperlinks
\hypersetup{
    colorlinks = true,
    citecolor = {blue},
    urlcolor = {black}}

\title{\LARGE \bf ET-Former: Efficient Triplane Deformable Attention for 3D Semantic Scene Completion From Monocular Camera
}

\author{Jing Liang$^{1}$, He Yin$^{2}$, Xuewei Qi$^{2}$, Jong Jin Park$^{2}$, Min Sun$^{2}$, Rajasimman Madhivanan$^{2}$, Dinesh Manocha$^{1}$
\thanks{$^{1}$ University of Maryland, College Park; 
 $^{2}$ Amazon Lab126. %\tt{\{heyinz, qixuewei,jongpark, minnsun, rajasimm\}@amazon.com}
}%
}

\begin{document}

\maketitle
\thispagestyle{empty}
\pagestyle{empty}

\begin{abstract}
We introduce ET-Former, a novel end-to-end algorithm for semantic scene completion using a single monocular camera. Our approach generates a semantic occupancy map from single RGB observation while simultaneously providing uncertainty estimates for semantic predictions. By designing a triplane-based deformable attention mechanism, our approach improves geometric understanding of the scene than other SOTA approaches and reduces noise in semantic predictions. Additionally, through the use of a Conditional Variational AutoEncoder (CVAE), we estimate the uncertainties of these predictions. The generated semantic and uncertainty maps will help formulate navigation strategies that facilitate safe and permissible decision making in the future. Evaluated on the Semantic-KITTI dataset, ET-Former achieves the highest Intersection over Union (IoU) and mean IoU (mIoU) scores while maintaining the lowest GPU memory usage, surpassing state-of-the-art (SOTA) methods. It improves the SOTA scores of IoU from $44.71$ to $51.49$ and mIoU from $15.04$ to $16.30$ on SeamnticKITTI test, with a notably low training memory consumption of $10.9$ GB, achieving at least a $25\%$ reduction compared to previous methods. Project page: \url{https://github.com/jingGM/ET-Former.git}.
\end{abstract}

\section{Introduction}

% what is semantic occupancy prediction, why is it important: applications
In robotics and autonomous driving, 3D scene understanding supports both navigation and interaction with the environment~\cite{saxena2008make3d, zhang2021holistic}. Semantic Scene Completion (SSC), also referred to as Semantic Occupancy Prediction, jointly predicts the semantic and geometric properties of an entire scene using a semantic occupancy map, including occluded regions and areas beyond the camera’s field of view (FOV)~\cite{li2023voxformer, zheng2024monoocc, cao2022monoscene}. While many approaches leverage depth cameras~\cite{choe2021volumefusion, dai2020sg} or LiDAR sensors~\cite{li2023lode, xia2023scpnet} for 3D environment perception, these sensors are often more expensive and less compact than mono-RGB cameras~\cite{cao2022monoscene}. As a result, mono-RGB cameras have gained increasing attention for 3D scene understanding in both autonomous driving and robotic navigation tasks~\cite{davison2007monoslam, li2023voxformer, zheng2024monoocc}. However, SSC using mono-cameras poses several challenges, such as accurately estimating semantics in the 3D space, predicting occluded areas outside the FOV, and get the robustness of the estimation.

% challenges of Occupancy prediction
Accurate estimation of pixel semantics and their real-world positions is difficult due to the inherent complexity of two key tasks~\cite{cao2022monoscene, zheng2024monoocc}: 1. semantic estimation that predicts semantic labels of the observed ares; and 2. precise projection from the image plane to the real-world environment that converts the 2D semantic areas into accurate 3D locations. From a single view, the predicted regions are strongly influenced by the pixel rays from the camera to the image plane~\cite{cao2022monoscene}, often resulting in skewed geometric shapes. In such cases, multi-view perception~\cite{li2024sgs, roldao20223d} or advanced feature processing~\cite{huang2023tri, zheng2024monoocc} can help correct these errors by incorporating the geometric structure of objects from different viewpoints.

However, estimating only the visible semantic voxels is insufficient to represent a complete 3D environment, as parts of the scene may be occluded or outside the camera’s FOV~\cite{huang2023tri, cao2022monoscene}. Estimating these occluded and out-of-FOV areas requires understanding the geometric structure and semantic relationships of nearby voxels~\cite{li2023voxformer, zheng2024monoocc}. Moreover, quantifying the uncertainty in these predictions is also crucial, as it provides critical guidance for navigation strategies of safety-critical applications, like autonomous driving and rescue missions. 

\begin{figure}
    \centering
    \includegraphics[width=0.8\linewidth]{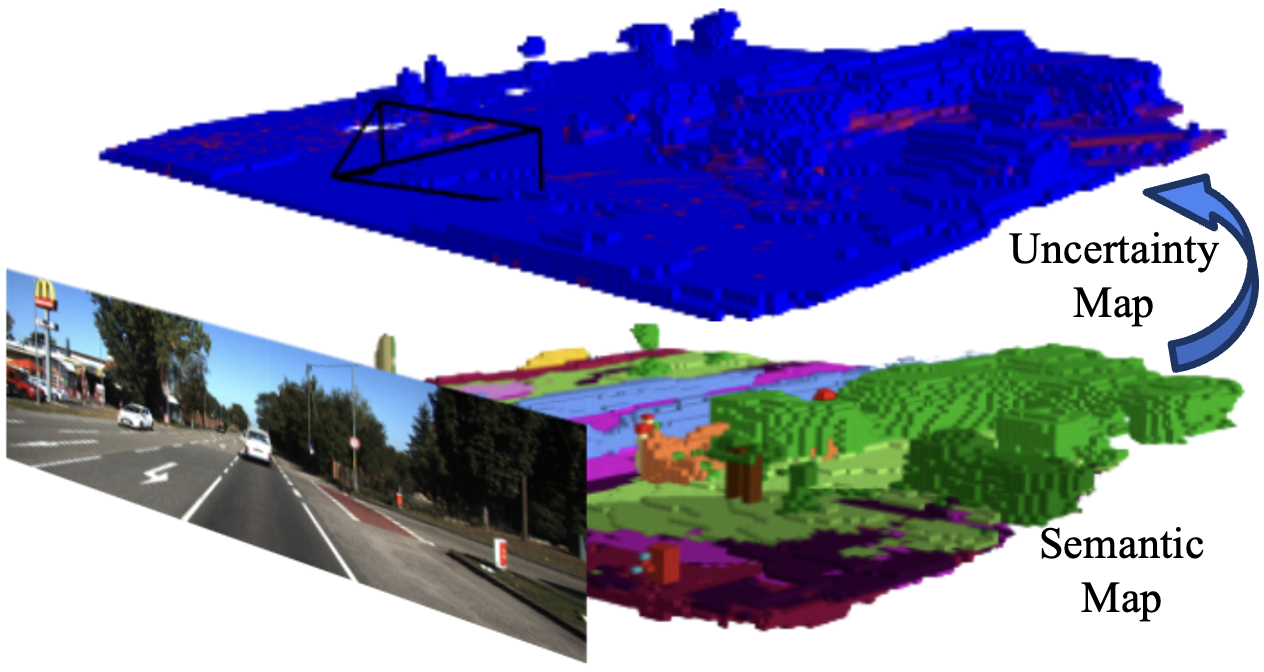}
    \caption{Our ET-Former predicts the semantic occupancy map from a monocular camera image to complete the scene and estimate the uncertainties of the occupancy map.}
    \label{fig:front}
                \vspace{-2em}
\end{figure}

\noindent\textbf{Main Contributions:} In this work, we address the challenge of accurate semantic estimation by proposing a triplane-based deformable attention model that uses deformable attention mechanism~\cite{zhu2020deformable, xia2022vision} to process voxel and RGB features in three orthogonal views. Our approach provides a better geometric understanding from three views than only mono-cam frame and our deformable attention also reduces the computational cost of feature processing. To predict occluded and out-of-view regions, we employ a Conditional Variational AutoEncoder (CVAE) to generate semantic voxel predictions and also estimate the uncertainties of the predictions. Our approach takes advantage of the two stages idea from VoxFormer~\cite{li2023voxformer}, where stage~1 estimates occupancy from the mono-camera, providing guidance to stage~2 for completing the 3D semantic occupancy map. Moreover, our triplane-based feature processing mechanism improves upon the deformable attention model in VoxFormer by smoothing estimation noise generated by sparse query voxels, and our CVAE generator performs more accurate prediction than other approaches. Our main contributions are as follows:
\begin{enumerate}
    \item We introduce a triplane deformable model that converts 2D image features into a 3D semantic occupancy map by projecting 3D voxel queries onto three orthogonal planes, where self- / cross-deformable attentions are applied. This model significantly reduces GPU memory usage of existing methods by at least $25\%$ while refining visual and geometric information from multiple views. Evaluated on the Semantic KITTI dataset, it improves the SOTA scores of IoU from $44.71$ to $51.49$ and mIoU from $15.04$ to $16.30$, with a notably low GPU memory consumption of $10.9$ GB.
    \item We designed an efficient cross deformable attention model for multiple source feature processing, such as 2D image features with 3D voxel features. The deformable attention model reduces the computational cost by using flexible number of reference points in the deformable attention mechanism.
    \item We adapt the formulation of CVAE, treating voxel queries as latent variables and reparameterizing them as Gaussian distributions to quantify uncertainty in the predicted occupancy map from stage 1, producing an uncertainty map. The integration of uncertainty and semantic maps will enable design of both safe and permissible navigation strategies in the future.
    \item We apply the triplane deformable model to stage 1 by fully leveraging image features to improve occupancy prediction accuracy. 
\end{enumerate}

% \begin{enumerate}
%     \item 9/3: video
%     \item 9/5: paper
%     \item 9/14: test real-world video
% \end{enumerate}

\section{Related Works}
\textbf{Camera-based Scene Understanding: }
3D scene understanding using cameras requires a comprehensive understanding of both geometric and color information~\cite{han2019image, li2022bevformer, huang2021bevdet}, such as 3D scene reconstruction~\cite{popov2020corenet, li2023voxformer}, 3D object detection~\cite{huang2023tri, huang2021bevdet}, and 3D depth estimation~\cite{zhang2022dino, lin2017feature}. For tasks like detection and depth estimation, pixels and points are widely utilized to represent 3D objects~\cite{huang2021bevdet, huang2023tri, zhang2022dino, lin2017feature}. In scene reconstruction and completion tasks, early methods applied Truncated Signed Distance Functions (TSDFs) to process features~\cite{dai2020sg, park2019deepsdf}. Then voxel-based approaches gained more popularity due to their sparsity and memory efficiency~\cite{cao2022monoscene, li2023voxformer}. However, converting RGB features to 3D voxel features remains challenging because it requires both color and geometric understanding of the environment. To address this, Bird's-Eye-View (BEV) scene representations are commonly used~\cite{li2022bevformer, zhang2022beverse}. The Lift-Splat-Shoot (LSS) method~\cite{philion2020lift} projects image features into 3D space using pixel-wise depth to generate BEV features. BEV-based methods like BevFormer~\cite{li2022bevformer} and PolarFormer~\cite{jiang2023polarformer} leverage attention mechanisms to learn the correlation between image features and 3D voxel features. These methods explicitly construct BEV features, while other works implicitly process BEV features within the network~\cite{reiser2021kilonerf, chen2021learning, huang2023tri}, demonstrating higher computational efficiency and the ability to encode arbitrary scene resolutions. Recently, triplane features have been introduced to extend BEV representations~\cite{huang2023tri, shue20233d}, projecting point features into three orthogonal planes instead of a single BEV plane. This approach has shown improved accuracy in 3D object detection~\cite{huang2023tri, huang2024gaussianformer}. However, these triplane-based approaches process do not process the voxels in the entire 3D scenarios. In our approach, we designed an efficient deformable-attention model to address the issue.

% SSC
\textbf{Semantic Scene Completion: } 
Semantic Scene Completion (SSC) involves estimating both the geometry and semantic labels of a 3D scene, a task first introduced by SSCNet~\cite{li2023sscnet}. Early methods~\cite{chen20203d, zhou2022cross, song2017semantic, reading2021categorical} utilized geometric information to correlate 2D image features with 3D space. MonoScene~\cite{cao2022monoscene} and LMSCNet~\cite{roldao2020lmscnet} with monocular cameras proposed end-to-end solutions that convert RGB images into 3D semantic occupancy maps through convolutional neural networks (CNNs) or U-Net architectures. TPVFormer~\cite{huang2023tri} applied triplane features for pixel processing with multiple cameras but struggled with filling in missing areas effectively and heavy memory usages. More recent approaches like OccFormer~\cite{zhang2023occformer} and VoxFormer~\cite{li2023voxformer} have utilized transformers to process features of monocular cameras. VoxFormer decouples occupancy and semantic predictions into two stages to improve accuracy in both semantic estimation and occlusion prediction. Building on VoxFormer, MonoOcc~\cite{zheng2024monoocc} and Symphonies~\cite{jiang2024symphonize} introduced more complex structures for both stages, further enhancing occupancy prediction and semantic estimation, though at the expense of higher memory usage. However, these methods do not estimate the certainty of semantic predictions, limiting their effectiveness in downstream tasks. To address this, we design a triplane-based deformable attention model to improve the estimation accuracy and incorporate Conditional Variational AutoEncoder (CVAE) method to predict the uncertainty of semantic estimations.

\begin{figure*}
    \centering
    \includegraphics[width=0.9\linewidth]{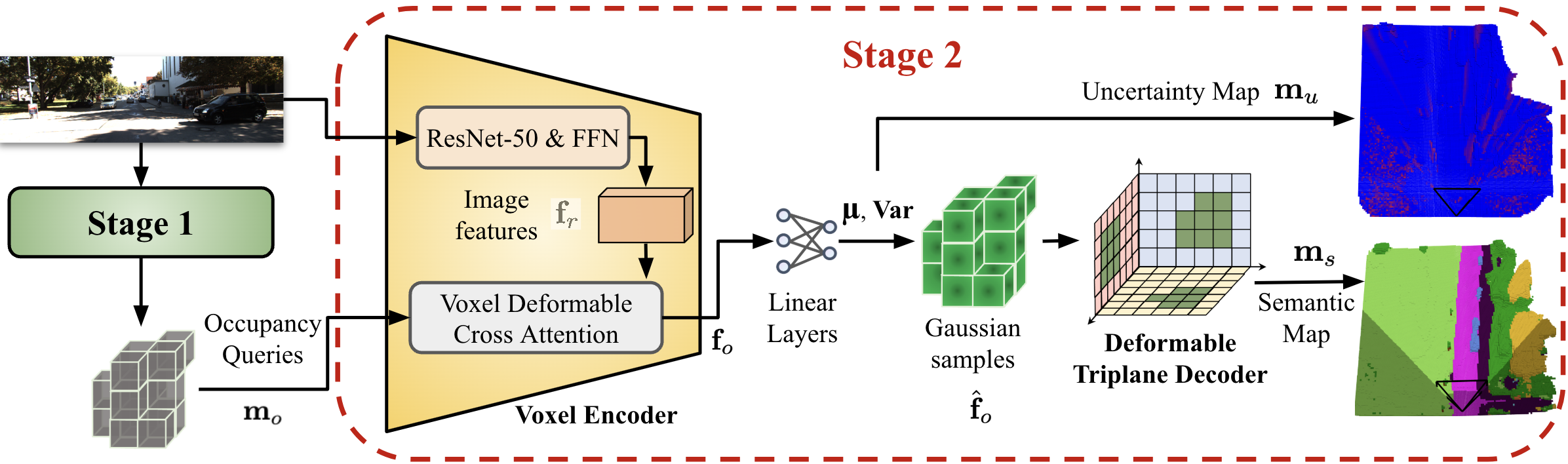}
    \caption{\textbf{Overall Architecture}: We present a two-stage pipeline for processing mono-cam images and generate both a semantic occupancy map $\m_s$ and its corresponding uncertainty map $\m_u$. In stage~1, we introduce a novel triplane-based deformable attention model to generate the occupancy queries $\m_o$ from the given mono-cam images, which reduces high-dimensional 3D feature processing to 2D computations. The detailed artchitecture of stage~1 is shown in Figure~\ref{fig:stage1}. In stage~2, we employ the efficient triplane-based deformable attention mechanism (detailed in Figure~\ref{fig:stage2}) to generate the semantic map, with the inferred voxels from stage~1 as input and conditioned on the RGB image. To estimate the uncertainty in the semantic map, we incorporate a CVAE method, and quantify the uncertainty using the variance of the CVAE latent samples.}
    \label{fig:architecture}
                \vspace{-2em}
\end{figure*}

\section{Our Approach}
In this section, we introduce the problem definition, key innovations including the efficient triplane-based deformable 3D attention mechanism, and the Conditional Variational Autoencoder (CVAE) formulation.

\subsection{Problem Definition}
We formulate the Semantic Scene Completion (SSC) task as a generative problem. Given a monocular RGB image $\i\in\RR^{H \times W \times 3}$, where $(H, W)$ represents the image size, our model aims to generate a semantic occupancy map $\M\in \RR^{L\times V\times D\times N}$, where $(L, V, D)$ represents the dimensions of the 3D occupancy map and $N$ is the number of semantic classes. The model follows a CVAE formulation:
\begin{align}
    p(\m_s|\i) &= \int_\z p(\m_s|\z,\i) p(\z)d\z, \;\;\;\z \sim p_\theta(\z|\i),
\end{align}
where $\m_s \in \M$ is the generated semantic occupancy map given the image instance $\i$. $p(\m_s|\z,\i)$ denotes the conditional distribution of generating the semantic map $\m_s$ given the latent sample $\z$ and image $\i$. $p(\z)$ is the prior distribution of $\z$. $p_\theta(\z|\i)$ represents the approximate posterior distribution of the latent representation $\z$, given the image $\i$. 

As illustrated in Figure~\ref{fig:architecture}, the model employs of a two-stage architecture for the SSC task. In stage~1, the occupancy map (queries) $\m_o\in \RR^{L\times V\times D\times 2}$ is estimated using the given image. The inferred $\m_o$ serves as the input for semantic prediction of stage~2. In stage~2, the semantic occupancy map $\m_s$ is generated using the CVAE formulation, with the inferred voxels from stage~1 as input and conditioned on the RGB image. As a byproduct of the CVAE formulation, reparameterizing the latent variables as Gaussian distributions allows for quantification of uncertainty in the occupancy map $\m_o$ from stage~1, giving rise to an uncertainty map $\m_u$.

% The Encoder model of stage two utilizes the ResNet-50~\cite{he2016deep} and FPN~\cite{lin2017feature}, which show high potential in image feature processing in vision tasks, to process the RGB image features into a smaller dimension vector $\f_r\in \RR^{w,h,d}$, where $w<W$, $h<H$, and $d$ indicates the feature dimension. Similar to VoxFormer~\cite{li2023voxformer}, DETR~\cite{zhu2020deformable} is used to enhance the occupancy queries with image features, where the feature map is $\f_r$ and the query is $\m_o$. The enhanced query feature $\f_o$ is calculated as in Equation~\ref{eq:detr_cross}

One of our major innovations is the Deformable Triplane Decoder, which is primarily designed for stage~2 but also serves as the backbone of stage~1 to provide more visual information for occupancy prediction. Given this interconnected design, we will first describe the methodology of stage~2 in detail. Subsequently, we will elaborate on the sepecifics of stage~1 in Section~\ref{sec:stage1}, highlighting how the Deformable Triplane Decoder is adapted and utilized in this initial stage. 
\subsection{Voxel Encoder}
Our voxel encoder in stage~2 employs ResNet-50~\cite{he2016deep} and Feed Forward Network (FFN)~\cite{lin2017feature}, which have demonstrated high potential in image feature extraction for vision tasks~\cite{li2023voxformer, zheng2024monoocc, zhang2022dino}, to process the RGB image into lower-dimensional features $\f_r\in \RR^{w,h,d}$, where $w<W$, $h<H$, and $d$ denotes the feature dimension. DETR~\cite{zhu2020deformable} is used to enhance occupancy queries $\m_o$ by incorporating the image features $\f_r$ and outputs occupancy features $\f_o$, which is shown in Equation~\ref{eq:detr_cross}. This process is referred to as voxel deformable cross attention in Figure~\ref{fig:architecture}.
\begin{align}
    \f_o = DETR(\z_q, \hat{\p}_o, \f_r) = & \nonumber  \\
    \sum_{m=1}^M \W_m [ \sum_{k=1}^K A_{mqk}& \W'_m \f_r(\phi_s(\hat{\p}_o) + \phi_o(\z_q))],
    \label{eq:detr_cross}
\end{align}
where $\p_o$ represents the 3D position of each query voxel in $\m_o$ and $\hat{\p}_o$ indicats its normalized position. $\z_q = \p_e(\p_o)$ is the positional embedding of the query positions, and $\p_e()$ is a positional embedding function. $M$ and $K$ indicate the head number and the offset number. $\phi_s$ scales the normalized position $\hat{\p}_o$ into the image feature size $h\times w$, and $\phi_o$ calculates the positional offset with $\z_q$. $\f_r(\phi_s(\hat{\p}_o) + \phi_o(\z_q))$ is the feature at location $(\phi_s(\hat{\p}_o) + \phi_o(\z_q))$. $\W_m$ and $\W'_m$ are linear weights. $A_{mqk}=\phi_a (\z_q)$ is also computed from $\z_q$, where $\phi_a()$ represents Linear layers.

\subsection{Conditional Variational Autoencoder Formulation}
% \TODO{Expand a little bit more on this, since this is one of the major contributions.}
% Our approach use Conditional Variational AutoEncoder (CVAE) to generate semantic occupancy map, considering the unique properties~\cite{bao2017cvae, harvey2021conditional, sohn2015learning} of CVAE method: 1. CVAE allows for conditional generation, which generates the voxels satisfying the conditional information. We use CVAE to generate the semantic map and satisfying the RGB information of the monocular observation. 2. CVAE learns a latent space that is conditioned on specific attributes. This latent space is usually smooth and continuous, which means that small changes in the latent space can lead to meaningful variations in the generated images. 3. CVAE models the uncertainty in the generated data. Unlike standard deterministic models, CVAE is probabilistic and can handle cases where there are multiple valid outputs for the same condition.

Our approach utilizes a Conditional Variational Autoencoder (CVAE) to generate semantic occupancy maps, leveraging the unique properties of the CVAE method~\cite{bao2017cvae, harvey2021conditional, sohn2015learning}: 1. CVAEs enable conditional generation, allowing for the generation of voxels that adhere to specific conditional information. We employ this capability to generate semantic maps that conform to the RGB information from monocular observations. 2. CVAEs learn a smooth and continuous latent space. This property ensures that small changes in the latent space lead to meaningful variations in the generated voxels, enhancing the expressiveness of our semantic maps. 3. CVAEs inherently model uncertainty in the generated data, accommodating cases with multiple valid outputs for the same condition, unlike deterministic models. This probabilistic nature allows our approach to handle uncertainty in monocular semantic occupancy mapping more effectively.

Given the query features $\f_o$, in the CVAE format, we need to convert the feature embedding to the embedding of Gaussian distributions, which are shown in Equation~\ref{eq:gaussian}
\begin{align}
\hat{\f}_o \sim \cn\{\mu = \l_m(\f_o),\;\;\; \log \text{var}= \l_v(\f_o)\},
    \label{eq:gaussian}
\end{align}
where $\l_m$ and $\l_v$ are linear layers. $\mu$ and $var$ are the mean and variance of the distribution. The variance of the Gaussian distribution represents the uncertainty in the uncertainty map $\m_u$, as shown in Figure \ref{fig:architecture}.

\begin{figure*}
\centering
\includegraphics[width=0.9\linewidth]{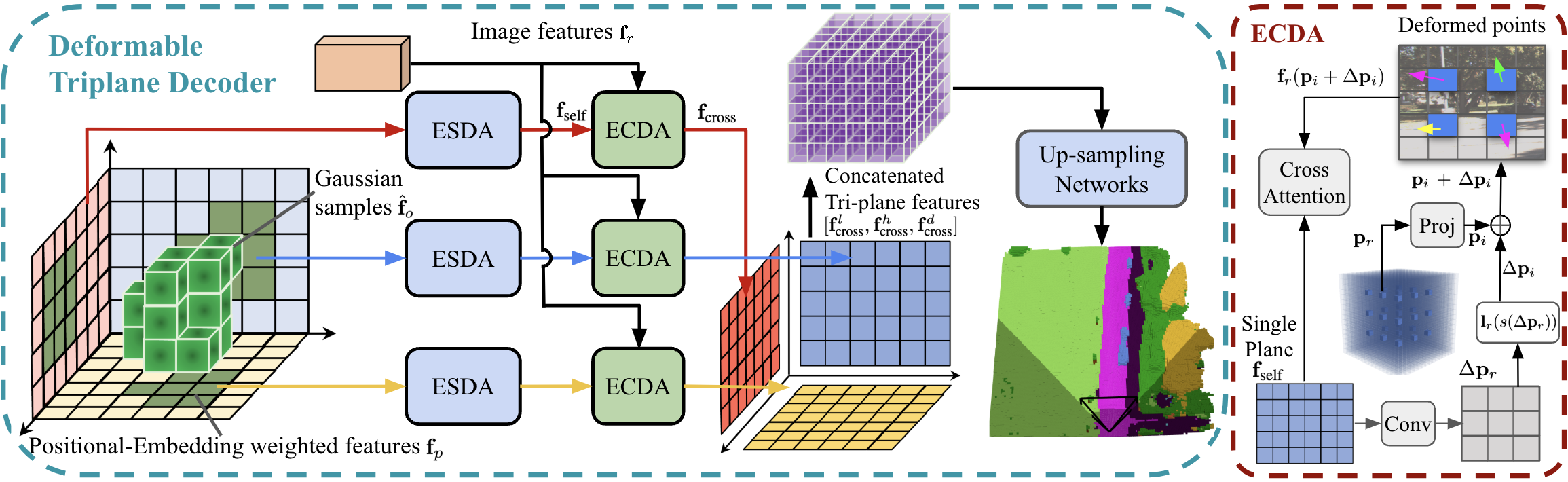}
    \caption{%\footnotesize
    \textbf{(a) Deformable Triplane Decoder of Stage 2}: We propose triplane-based efficient self-deformable attention (ESDA) and efficient cross-deformable attention (ECDA) methods in the decoder to generate 3D features. The input to the decoder consists of Gaussian samples $\hat{\f}_o$ and image features $\f_r$. The Gaussian samples are projected onto three orthogonal planes based on positional embeddings. The three planes features are completed by ESDA and then enhanced with image features, $\f_r$, by ECDA. Finally, the plane features are concatenated and upsampled to generate the semantic occupancy map. \textbf{(b) Detailed Structure of ECDA}: ECDA utilizes image features at deformed points $(\p_i + \Delta \p_i)$ and completed plane feature $\f_{\text{self}}$ to conduct cross attention.
    }
    \label{fig:stage2}
                \vspace{-1.5em}
\end{figure*}

\subsection{Triplane-based 3D Deformable Attention}
\label{sec:tda}
The triplane-based representation~\cite{huang2023tri, shue20233d} encodes 3D features by leveraging three orthogonal 2D feature planes ($XY$, $XZ$, and $YZ$). This approach significantly reduces both memory and computational requirements compared to directly encoding features in dense 3D grids. In comparison to bird's-eye-view representations, triplane encoding offers a more sufficient representation of the real-world environment. The orthogonality of the three planes provide a robust embedding of spatial features within the 3D space, making it an effective and efficient representation for complex 3D scenes. 
% Deformable attention mechanism~\cite{zhu2020deformable, xia2022vision} shows effiency in processing high-resolution features with spatial awareness of the image pixels. However, the current methods still use very large query and reference points in calculations, and the computational cost is still very high. We propose an efficient triplane-based deformable attention method to process cross-source features (image and voxel features), where the image features are used to enhance the voxel features.
The deformable attention mechanism~\cite{zhu2020deformable, xia2022vision} is efficient in processing high-resolution features with spatial awareness of image pixels. However, current methods~\cite{zhu2020deformable, li2023voxformer} still rely on large sets of query and reference points, resulting in high computational costs for 3D feature processing. To address this, we propose an efficient triplane-based deformable attention method that enhances voxel features with image features $\f_r$, improving the efficiency and performance of cross-source feature extraction. % To address this, we propose an efficient triplane-based deformable attention method to process cross-source features (image and voxel features), where the image features are used to enhance the voxel features, improving the overall efficiency and performance of feature extraction across different data modalities. 

We propose an efficient deformable triplane  decoder in stage~2. Instead of querying the entire 3D voxel space, we reduce the dimensionality from 3D to 2D by aggregating voxel features onto three orthogonal planes. We then propose efficient self- and cross-deformable attention mechanisms to process the features of these planes. As shown in Figure~\ref{fig:stage2}, the query features $\hat{\f}_o$ are aggregated by the positional embedding weights:
\begin{align}
    \f_p(x, y) = \sum_{z=0}^Z \p_e(\p_o(x,y,z)) \circ \hat{\f}_o(x,y,z),
    \label{eq:projection}
\end{align}
where $x$ and $y$ are the cell indices in a %$XY$
plane, which also correlate to the $x,y$ in the 3D space, and $z$ represents the third axis in the 3D space. $\p_e$ is a positional embedding function. $\hat{\f}_o$ is a sample from the normal distribution in Equation~\ref{eq:gaussian}. The notation $\circ$ in Equation~\ref{eq:projection} represents element-wise product. In short, $\f_p$ at location $(x, y)$ is the weighted sum of $\hat{\f}_o(x,y,z)$ across all values of $z$. The same aggregation process also applies to other two planes. 

Since queries projected from 3D space to 2D planes only occupy part of the 2D space, there are empty cells in the 2D space. Therefore, to have a full representation of the 3D space, we must fill in the missing cell features. As for the SemanicKITTI dataset \cite{behley2019semantickitti}, the missing cells are more than half of the plane,
% around $67\%$ of each plane, 
so the plane features are sparse. We employ the Deformable Transformer~\cite{xia2022vision} to complete the missing cell features, using plane-wide cell features as queries. The reference points are downsampled from the image feature space at a stride of four pixels to strike a balance between computation efficiency and performance. The Efficient Self-Deformable Attention (ESDA) mechanism completes the three planes, as $\f_\text{self} = \text{ESDA}(\f_p)$.

Since the decoder follows the CVAE format, we leverage image features, $\f_r$, as a condition to provide different view information and color information, thereby enhancing the features of the three orthogonal planes. We introduce a new Efficient 3D Cross Deformable Attention (ECDA) model to process the plane features $\f_\text{self}$ with the image feature $\f_r$, which is illustrated in Figure~\ref{fig:stage2}. Since the plane features are converted from 3D to 2D, we retain 3D geometric representation by uniformly sampling reference points $\p_r\in \RR^{l,v,d}$ from the 3D voxel grid, where $l\ll L$, $v\ll V$, and $d\ll D$. These points in the camera's view, can be projected to the image feature plane $\RR^{w\times h}$, so we have $N_r = l\times v\times d$ reference points $\p_i\in\RR^{l\times v\times d\times 2}$ in the image feature plane. The filled-in triplane features, $\f_\text{self}$, serve as queries in ECDA. Because the reference points are uniformly sampled from the 3D space, the $x, y$ coordinates are correlated to the triplane $x, y$ positions. We can use convolutional layers directly to calculate the offsets of the reference points, as $ \Delta \p_r = \text{Conv}(\f_\text{self})$, where $\Delta\p_r\in \RR^{l \times v \times c}$ and $c$ is the feature dimension. We separate the feature dimension into $d$ vectors to represent the offset of $d$ points in the $z$ axis. Finally, we have $\Delta \p_i = \l_r(s(\Delta\p_r))$, where $s()$ represents the feature separation, and $\l_r$ is a linear layer.
% $\Delta \p_i = \l_r(\text{rearrange} (\Delta\f_r))$. 
% The rearrange$(\cdot)$ is to decompose the dimension $c$ into $d\times c$ \textcolor{red}{what is rearrange here? Is that vector separation operation?} 
Then a multi-head cross attention model is used to process the sampled image features with the plane features as in Equation~\ref{eq:cross_attn}:
\begin{align}
    \f_\text{cross} = \text{CrossAttention} (\f_\text{self}, \f_r(\p_i+\Delta\p_i)),
    \label{eq:cross_attn}
\end{align}
% \begin{align}
%     \sum_{m=1}^M\W_m[\sum_{k=1}^K A_{mqk}\cdot W_m' \x(\p_i+\Delta\p_i)] \Rightarrow \text{MultiHead} (\z_q, \x(\p_i+\Delta\p_i))
% \end{align}
where $\f_\text{self}$ is the query. $\f_r(\p_i+\Delta\p_i)$, the image feature at deformed point locations $(\p_i+\Delta\p_i)$, serves as key and value of the multi-head attention model. The $\f_\text{cross}$ from three planes would be concatenated together and processed by several linear layers and upsampled to the semantic map, $\m_s = \l_s(\text{UpSample}(\l_c([\f_\text{cross}^l,\f_\text{cross}^h,\f_\text{cross}^d])))$, where $\l_s$ and $\l_c$ are all linear layers. $\f_\text{cross}^l$, $\f_\text{cross}^h$ and $\f_\text{cross}^d$ are features of the 2D planes orthogonal to the $l, h, d$ axes.

% \TODO{Draw picture of ECDA.} 

% \TODO{Might be great to briefly explain the difference between our triplane deformable decoder and TPVFormer} We are not based on any methods, so we dont' need to specifically compare with any method in algorithmic level.

\begin{table*}[h]
\scriptsize
    \centering
    \setlength{\tabcolsep}{3pt}
    \begin{tabular}{l|ll|lllllllllllllllllll}
       \hline
         Methods & IoU & mIoU &  \rotatebox{90}{\fcolorbox{white}{violet!40!magenta}{\makebox[0.1em]{\rule{0pt}{0.1em}}} road } & \rotatebox{90}{\fcolorbox{white}{violet!95!black}{\makebox[0.1em]{\rule{0pt}{0.1em}}} sidewalk} & \rotatebox{90}{\fcolorbox{white}{magenta!40}{\makebox[0.1em]{\rule{0pt}{0.1em}}} parking}&\rotatebox{90}{\fcolorbox{white}{red!40!brown}{\makebox[0.1em]{\rule{0pt}{0.1em}}} other-ground}&\rotatebox{90}{\fcolorbox{white}{yellow!90!black}{\makebox[0.1em]{\rule{0pt}{0.1em}}} building}&\rotatebox{90}{\fcolorbox{white}{blue!50!white}{\makebox[0.1em]{\rule{0pt}{0.1em}}} car}&\rotatebox{90}{\fcolorbox{white}{blue!70!red}{\makebox[0.1em]{\rule{0pt}{0.1em}}} truck}&\rotatebox{90}{\fcolorbox{white}{cyan!30}{\makebox[0.1em]{\rule{0pt}{0.1em}}} bicycle}&\rotatebox{90}{\fcolorbox{white}{blue!80!black}{\makebox[0.1em]{\rule{0pt}{0.1em}}} motorcycle}&\rotatebox{90}{\fcolorbox{white}{blue!70!white}{\makebox[0.1em]{\rule{0pt}{0.1em}}} other-vehicle}&\rotatebox{90}{\fcolorbox{white}{green!75!black}{\makebox[0.1em]{\rule{0pt}{0.1em}}} vegetation}&\rotatebox{90}{\fcolorbox{white}{brown!80!black}{\makebox[0.1em]{\rule{0pt}{0.1em}}} trunk}&\rotatebox{90}{\fcolorbox{white}{green!50!white}{\makebox[0.1em]{\rule{0pt}{0.1em}}} terrain}&\rotatebox{90}{\fcolorbox{white}{red!80!white}{\makebox[0.1em]{\rule{0pt}{0.1em}}} person}&\rotatebox{90}{\fcolorbox{white}{magenta!80}{\makebox[0.1em]{\rule{0pt}{0.1em}}} bicyclist}&\rotatebox{90}{\fcolorbox{white}{brown!60!violet}{\makebox[0.1em]{\rule{0pt}{0.1em}}} motorcyclist}&\rotatebox{90}{\fcolorbox{white}{orange!70!white}{\makebox[0.1em]{\rule{0pt}{0.1em}}} fence}&\rotatebox{90}{\fcolorbox{white}{yellow!30!white}{\makebox[0.1em]{\rule{0pt}{0.1em}}} pole}&\rotatebox{90}{\fcolorbox{white}{red!90!white}{\makebox[0.1em]{\rule{0pt}{0.1em}}} traffic-sign}\\
       \hline
% LMSCNet~\cite{roldao2020lmscnet}&31.38&7.07&46.70&19.50&13.50&3.10&10.30&14.30&0.30&0.00&0.00&000&10.80&0.00&10.40&0.00&0.00&0.00&5.40&0.00&0.00 \\ 
% AICNet~\cite{li2020anisotropic}    &23.93&7.09&39.30&18.30&19.80&1.60&9.60&15.30&0.70&0.00&0.00&0.00&9.60&1.90&13.50&0.00&0.00&0.00&5.00&0.10&0.00 \\ 
% JS3C-Net~\cite{yan2021sparse}   &34.00&8.97&47.30&21.70&19.90&2.80&12.70&20.10&0.80&0.00&0.00&4.10&14.20&3.10&12.40&0.00&0.20&0.20&8.70&1.90&0.30\\ 
MonoScene~\cite{cao2022monoscene} &34.16&11.08&54.70&27.10&24.80&5.70&14.40&18.80&3.30&0.50&0.70&4.40&14.90&2.40&19.50&1.00&1.40&0.40&11.10&3.30&2.10\\ 
TPVFormer~\cite{huang2023tri}      &34.25&11.26&55.10&27.20&27.40&6.50&14.80&19.20&3.70&1.00&0.50&2.30&13.90&2.60&20.40&1.10&2.40&0.30&11.00&2.90&1.50\\ 
VoxFormer~\cite{li2023voxformer}   &42.95&12.20&53.90&25.30&21.10&5.60&19.80&20.80&3.50&1.00&0.70&3.70&22.40&7.50&21.30&1.40&\textbf{2.60}&0.20&11.10&5.10&4.90\\ 
OccFormer~\cite{zhang2023occformer}&34.53&12.32&55.90&\textbf{30.30}&\textbf{31.50}& 6.50&15.70&21.60&1.20&1.50&1.70&3.20&16.80&3.90&21.30&2.20&1.10&0.20&11.90&3.80&3.70 \\ 
MonoOcc~\cite{zheng2024monoocc}&-&13.80&55.20&27.80&25.10&9.70&21.40&23.20&5.20&2.20&1.50&5.40&24.00&8.70&23.00&1.70&2.00&0.20&13.40&5.80&6.40 \\ 
Symphonies~\cite{jiang2024symphonize}&42.19&15.04&\textbf{58.40}&29.30&26.90&\textbf{11.70}&24.70&23.60&3.20&\textbf{3.60}&\textbf{2.60}&5.60&24.20&10.00&23.10&\textbf{3.20}&1.90&\textbf{2.00}&\textbf{16.10}&7.70&\textbf{8.00} \\ 
OctreeOcc~\cite{lu2023octreeocc}&44.71&13.12&55.13&26.74&18.68& 0.65&18.69&28.07&\textbf{16.43}&0.64&0.71&6.03&25.26&4.89&32.47&2.25&2.57&0.00&4.01&3.72&2.36 \\ \hline
%  ET-Former (stage 2)      & \textbf{43.92} & \textbf{13.05} & \textbf{57.78} & 25.45&18.68 &0.09 &18.61 &\textbf{28.19} & \textbf{13.64} &0.32& 0.88 &\textbf{5.43} & \textbf{25.22}&4.20 &\textbf{32.70} & 0.51& 0.80& 0.00&6.97 & \textbf{6.39}& 2.20 \\ 
ET-Former (Ours)      & \textbf{51.49} & \textbf{16.30} & 57.64 & 25.80 &16.68 &0.87 &\textbf{26.74} &\textbf{36.16} & 12.95 &0.69& 0.32 &\textbf{8.41} & \textbf{33.95} & \textbf{11.58} &\textbf{37.01} & 1.33&2.58& 0.32&9.52 &\textbf{19.60}&6.90 \\ 
       \hline
    \end{tabular}
    \caption{%\footnotesize
    \textbf{Quantitative Comparison On SemanticKITTI \texttt{test}  \cite{behley2019semantickitti}}: Our ET-Former outperforms other camera-based approaches, improving the SOTA scores of IoU from $44.71$ to $51.49$ and mIoU from $15.04$ to $16.30$. For specific semantic categories, such as building, car, other-vehicle, vegetation, trunk, terrain, and pole, our model provides the best accuracy.}
    \label{tab:ssc}
    % \vspace{-2em}
\end{table*}
\begin{figure*}[!ht]
  \centering
  \begin{tabular}{  c  c  c  c  c }
    % \hline
   % \multicolumn{3}{|c}{\textbf{Traversability Analysis}} & \multicolumn{3}{|c|}{\textbf{Heuristic Analysis}} \\ \hline
   \textbf{RGB} & \textbf{Ground Truth} & \textbf{ET-Former} & \textbf{VoxFormer}~\cite{li2023voxformer} & \textbf{MonoScene}~\cite{cao2022monoscene} \\ %\hline
   
    \begin{minipage}{.19\linewidth} \includegraphics[width=1.2\linewidth]{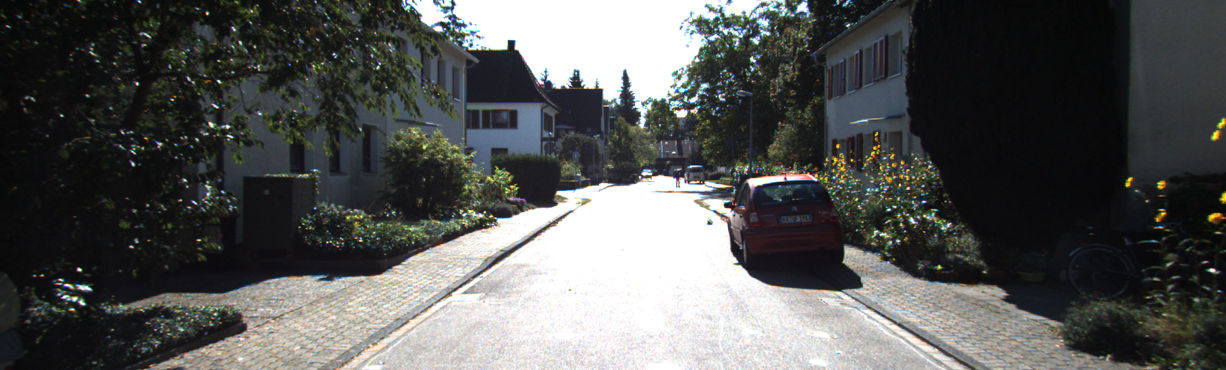} \end{minipage}
    &
    \begin{minipage}{.17\linewidth} \includegraphics[width=\linewidth]{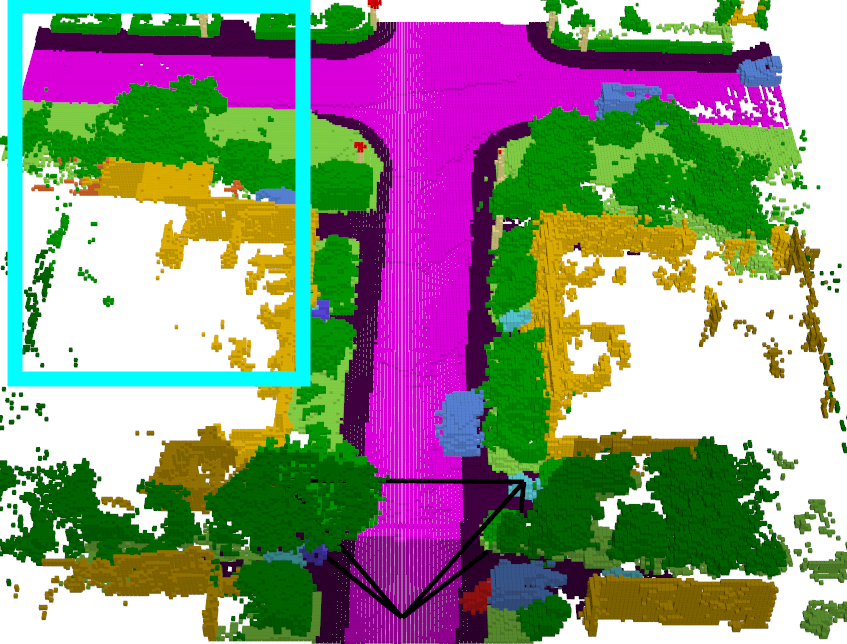} \end{minipage}
    &
    \begin{minipage}{.17\linewidth} \includegraphics[width=\linewidth]{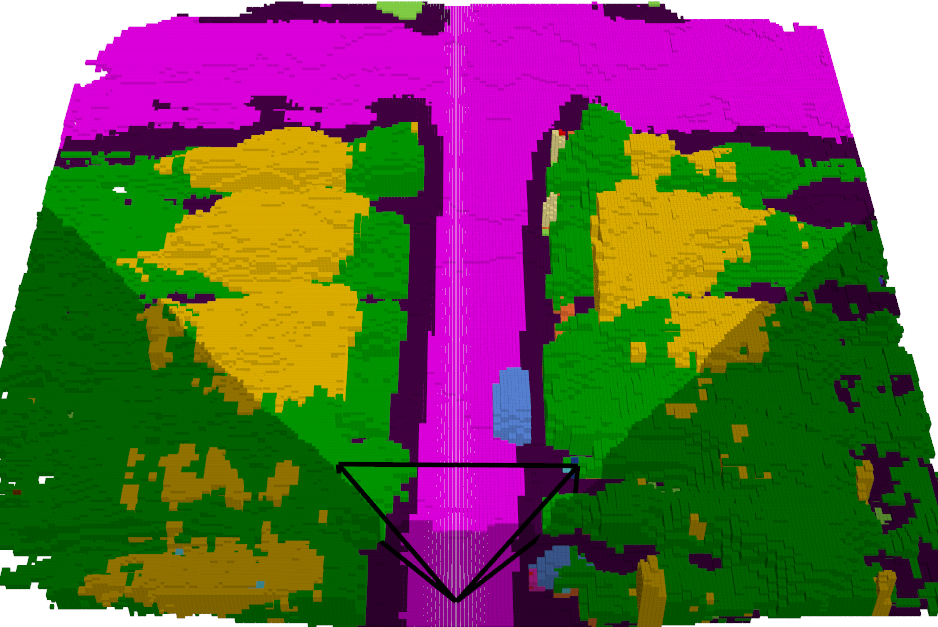} \end{minipage}
    &
    \begin{minipage}{.17\linewidth} \includegraphics[width=\linewidth]{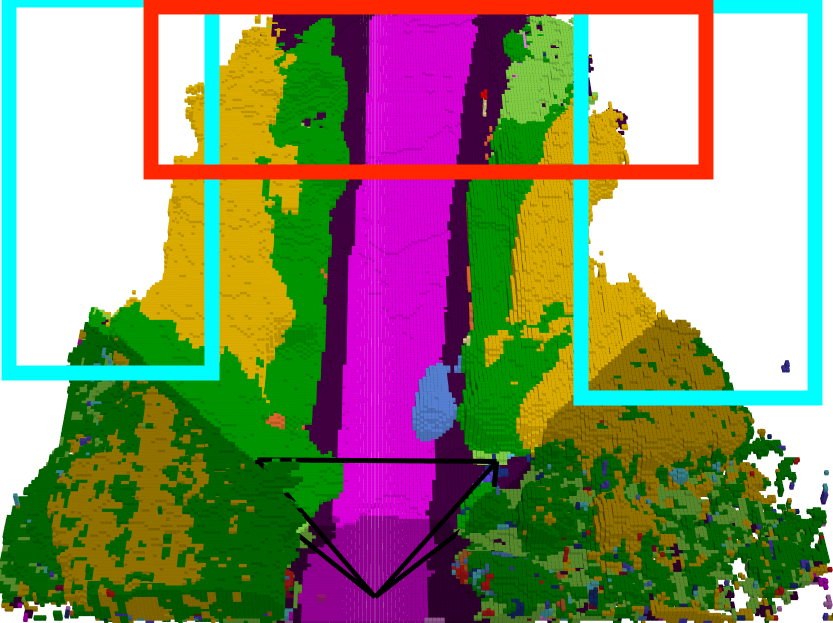} \end{minipage}
    &
    \begin{minipage}{.17\linewidth} \includegraphics[width=\linewidth]{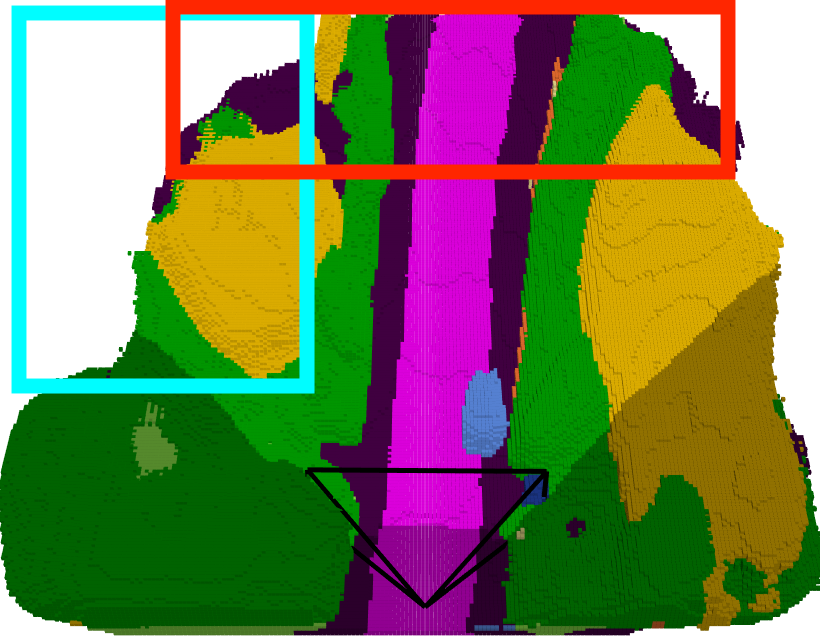} \end{minipage}
    \\ %\hline

    \begin{minipage}{.19\linewidth} \includegraphics[width=1.2\linewidth]{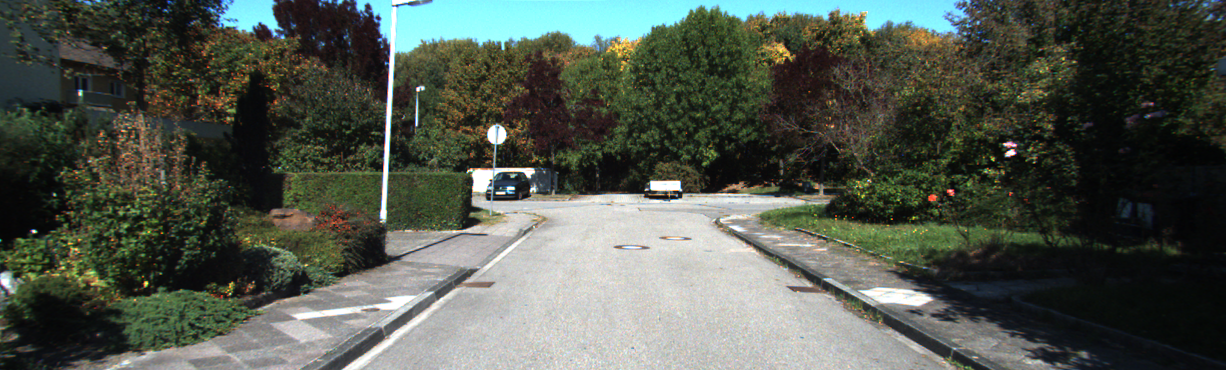} \end{minipage}
    &
    \begin{minipage}{.17\linewidth} \includegraphics[width=\linewidth,height=0.8\linewidth]{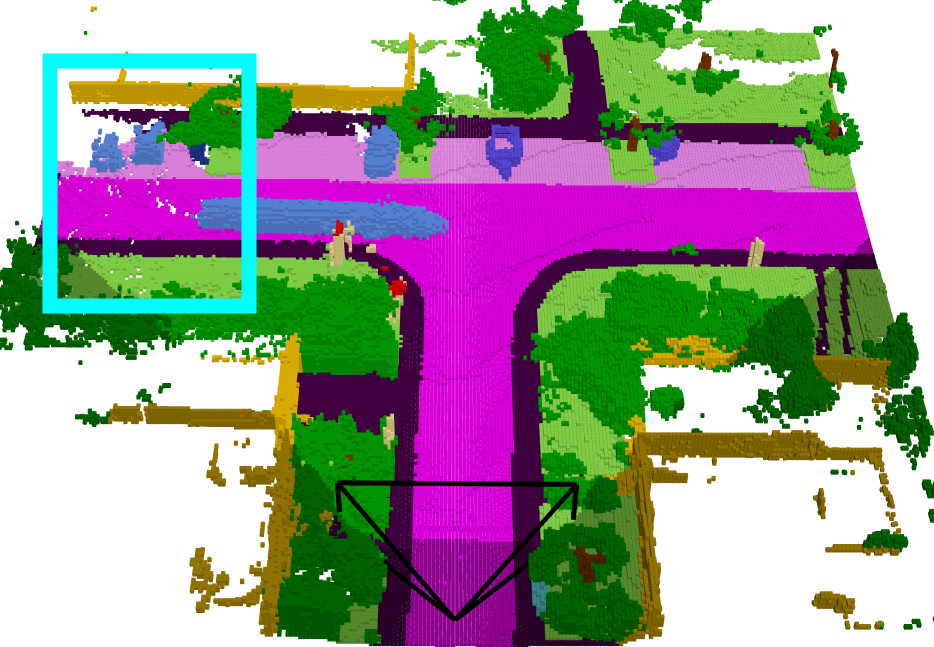} \end{minipage}
    &
    \begin{minipage}{.17\linewidth} \includegraphics[width=\linewidth]{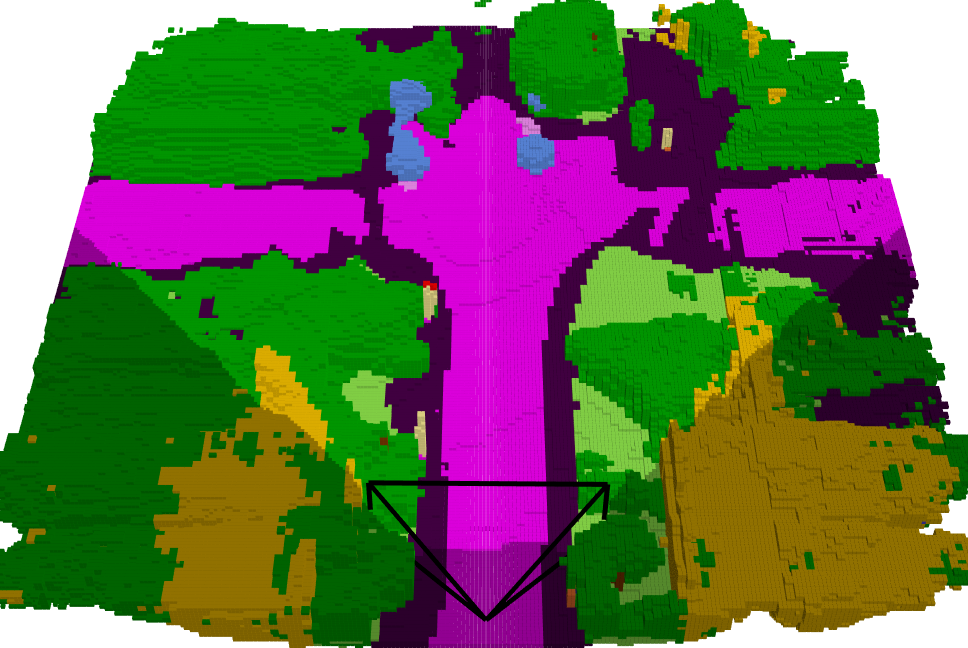} \end{minipage}
    &
    \begin{minipage}{.17\linewidth} \includegraphics[width=\linewidth]{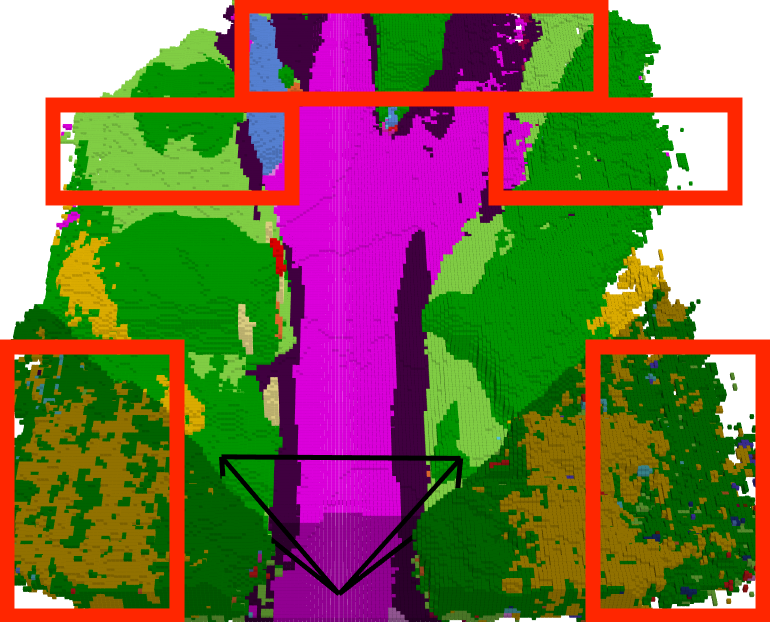} \end{minipage}
    &
    \begin{minipage}{.17\linewidth} \includegraphics[width=\linewidth]{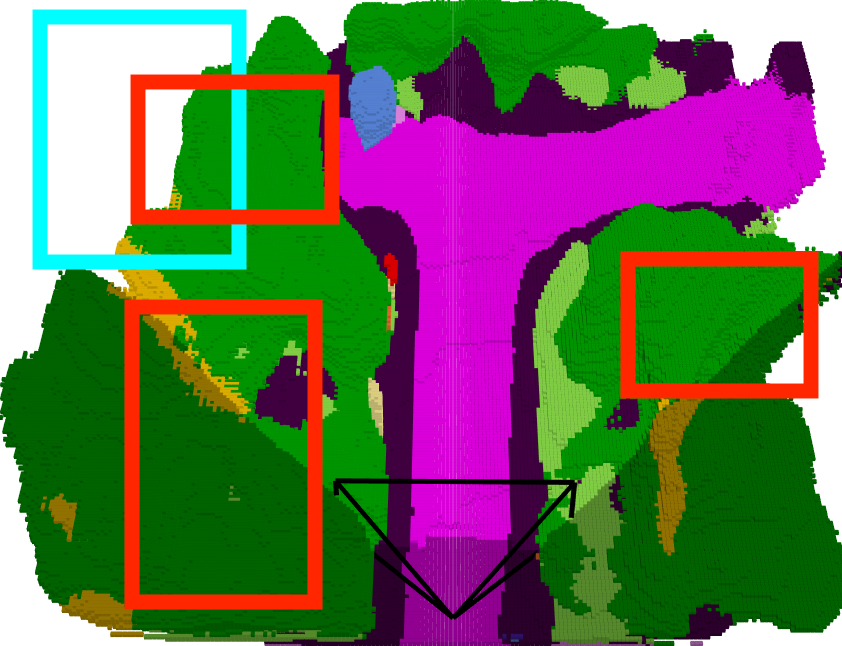} \end{minipage}
    \\ %\hline
    
    \begin{minipage}{.19\linewidth} \includegraphics[width=1.2\linewidth]{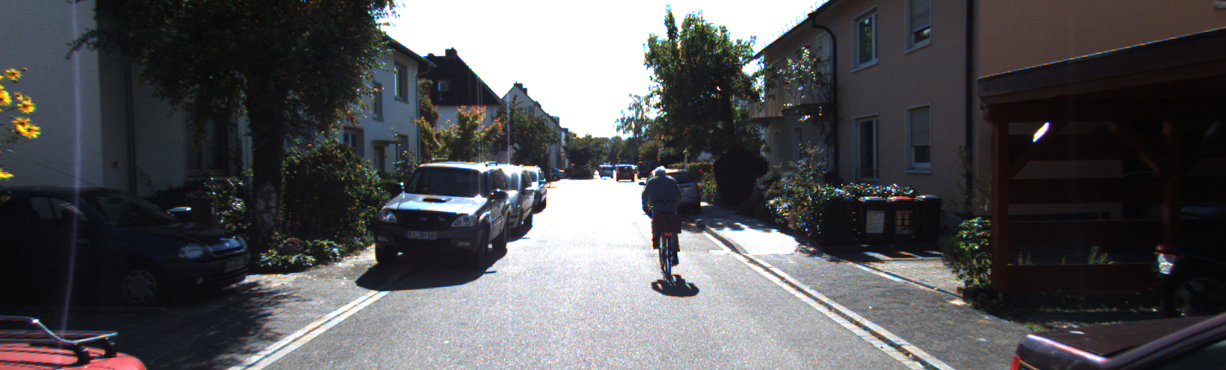} \end{minipage}
    &
    \begin{minipage}{.17\linewidth} \includegraphics[width=\linewidth,height=0.8\linewidth]{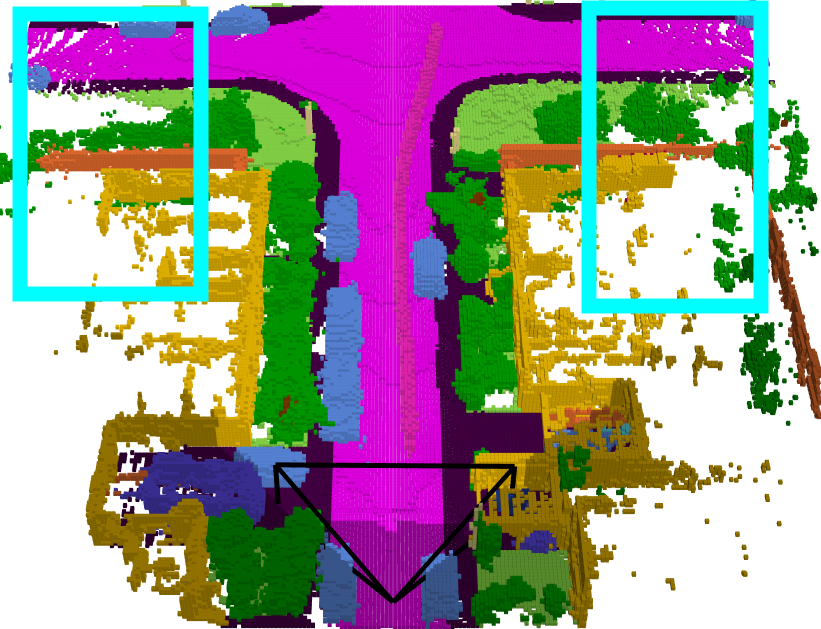} \end{minipage}
    &
    \begin{minipage}{.17\linewidth} \includegraphics[width=\linewidth]{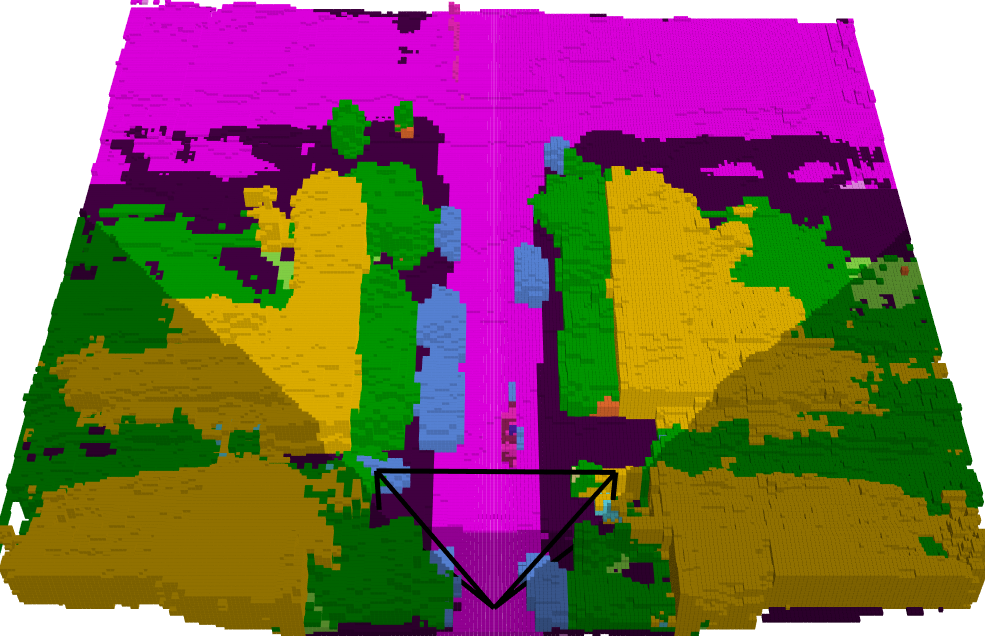} \end{minipage}
    &
    \begin{minipage}{.17\linewidth} \includegraphics[width=\linewidth]{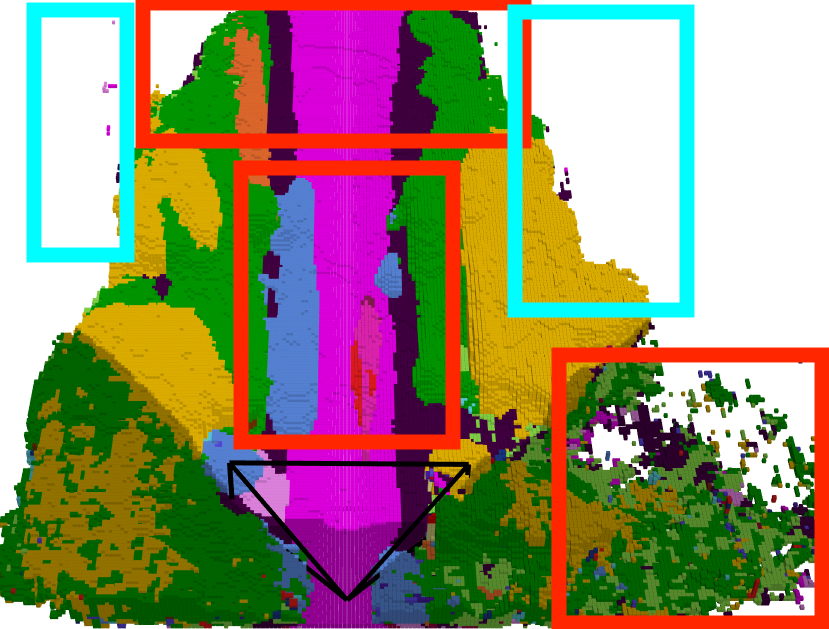} \end{minipage}
    &
    \begin{minipage}{.17\linewidth} \includegraphics[width=\linewidth]{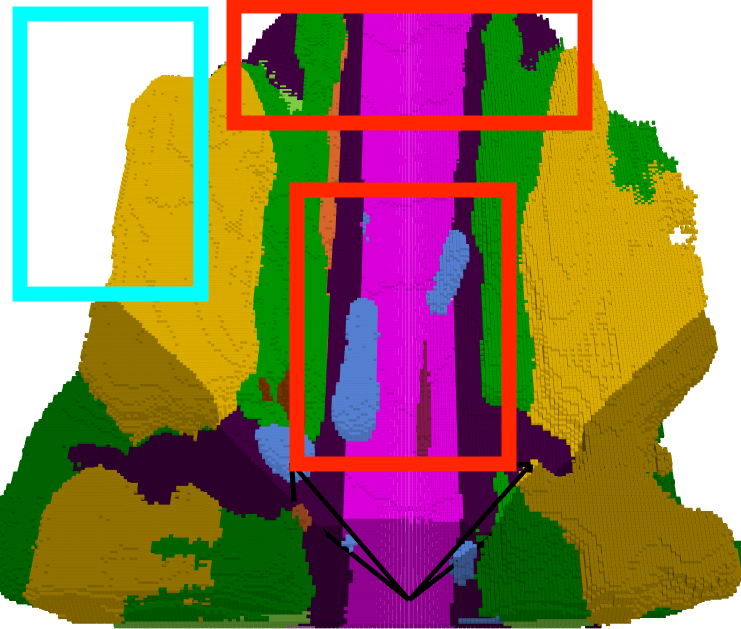} \end{minipage}
    \\ %\hline
  \end{tabular}
  \caption{%\footnotesize
  \textbf{Semantic Qualitative Results:} The blue rectangles highlight areas that are present in the Ground Truth but are missing from both VoxFormer and MonoScene predictions. The red rectangles marked the areas with wrong semantic estimations from Voxformer and MonoScene. }
  \label{fig:semantic_qualitative}
  % \vspace{-1.0em}
\end{figure*}

\begin{figure*}[!ht]
  \centering
  \begin{tabular}{  c  c  c  c }
    % \hline
   % \multicolumn{3}{|c}{\textbf{Traversability Analysis}} & \multicolumn{3}{|c|}{\textbf{Heuristic Analysis}} \\ \hline
   \textbf{RGB} & \textbf{Ground Truth} & \textbf{ET-Former} & \textbf{Uncertainty} \\ %\hline
   
    \begin{minipage}{.27\linewidth} \includegraphics[width=\linewidth]{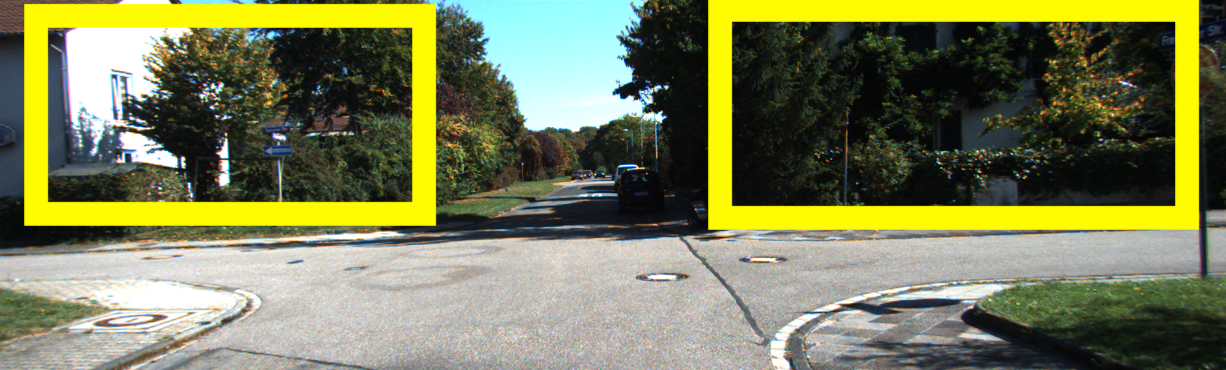} \end{minipage}
    &
    \begin{minipage}{.2\linewidth} \includegraphics[width=\linewidth]{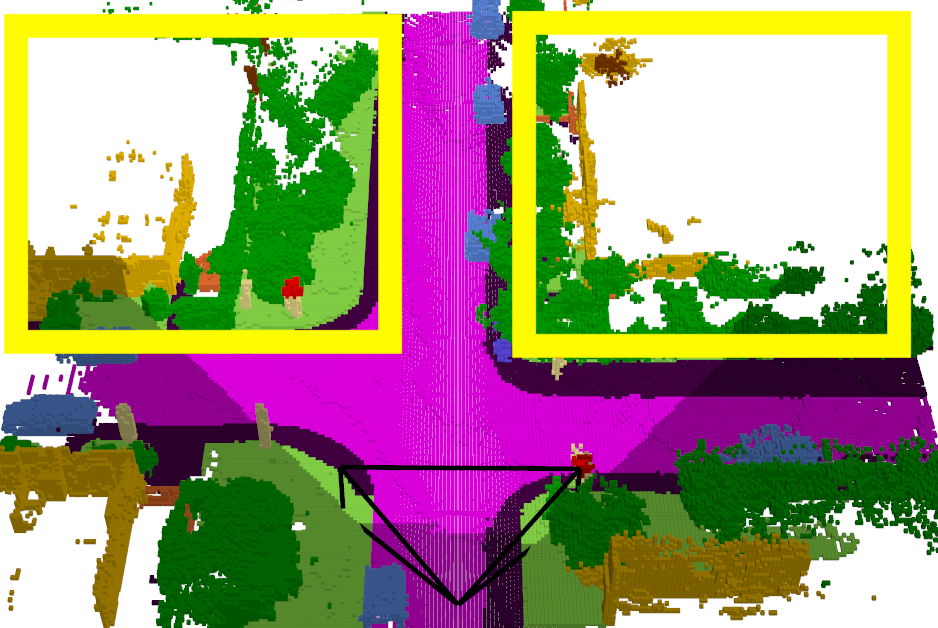} \end{minipage}
    &
    \begin{minipage}{.2\linewidth} \includegraphics[width=\linewidth]{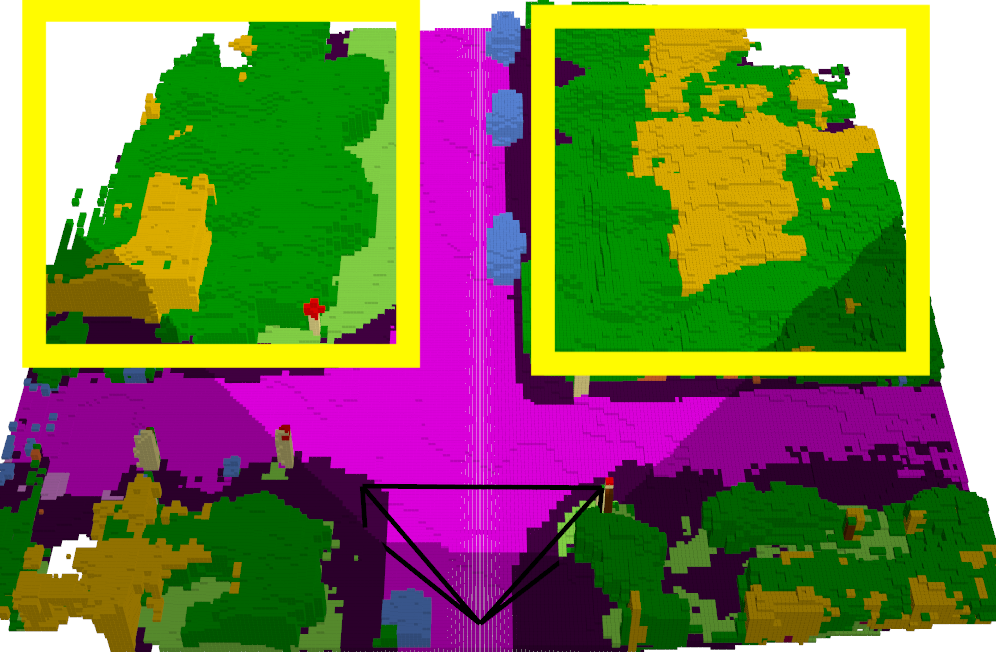} \end{minipage}
    &
    \begin{minipage}{.2\linewidth} \includegraphics[width=\linewidth]{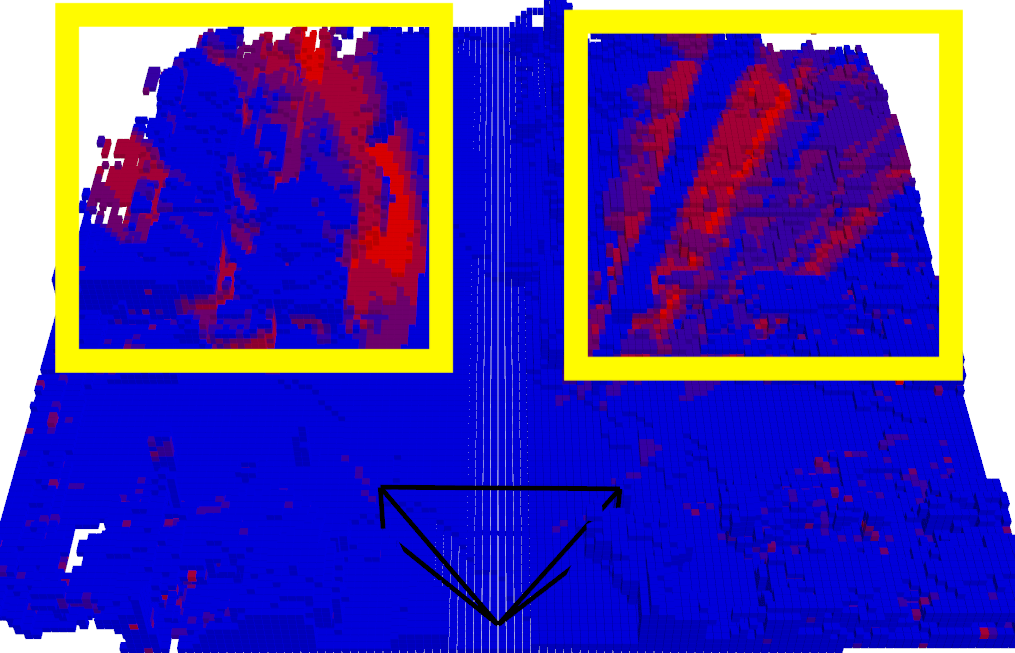} \end{minipage}
    \\ %\hline

    \begin{minipage}{.27\linewidth} \includegraphics[width=\linewidth]{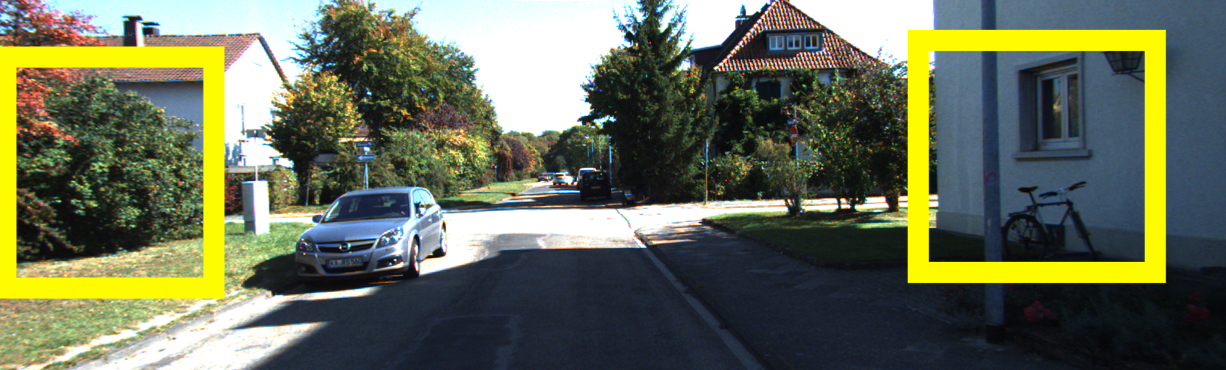} \end{minipage}
    &
    \begin{minipage}{.2\linewidth} \includegraphics[width=\linewidth]{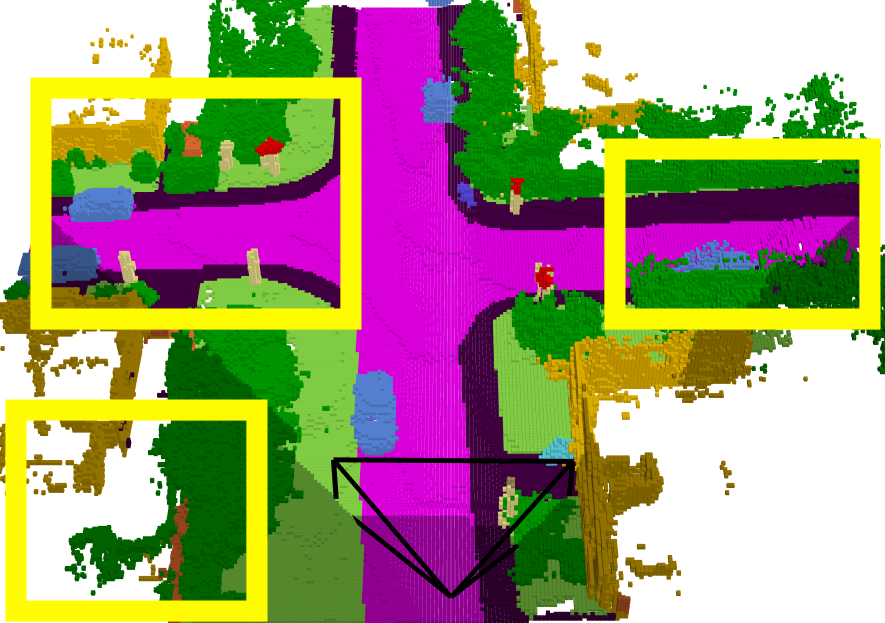} \end{minipage}
    &
    \begin{minipage}{.2\linewidth} \includegraphics[width=\linewidth]{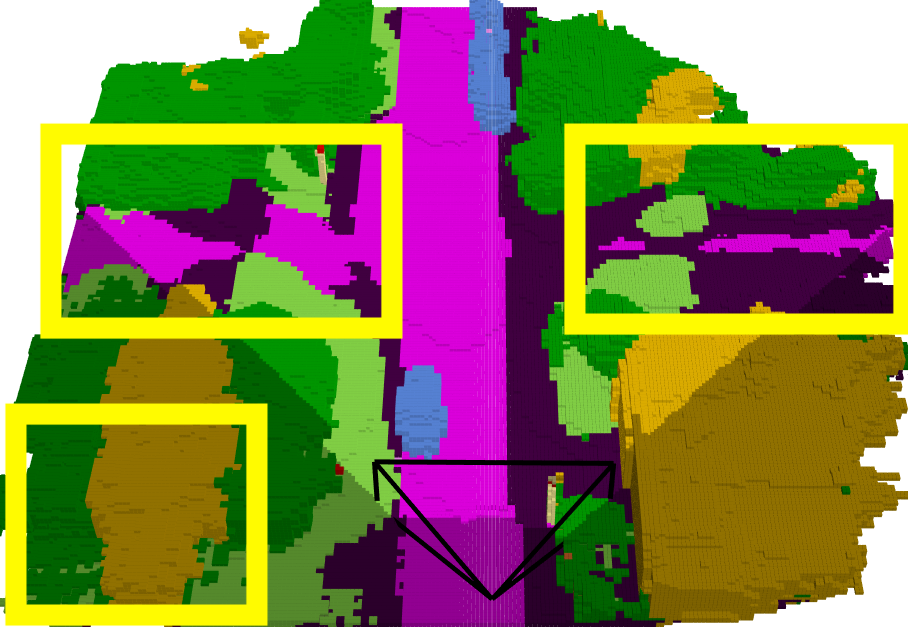} \end{minipage}
    &
    \begin{minipage}{.2\linewidth} \includegraphics[width=\linewidth]{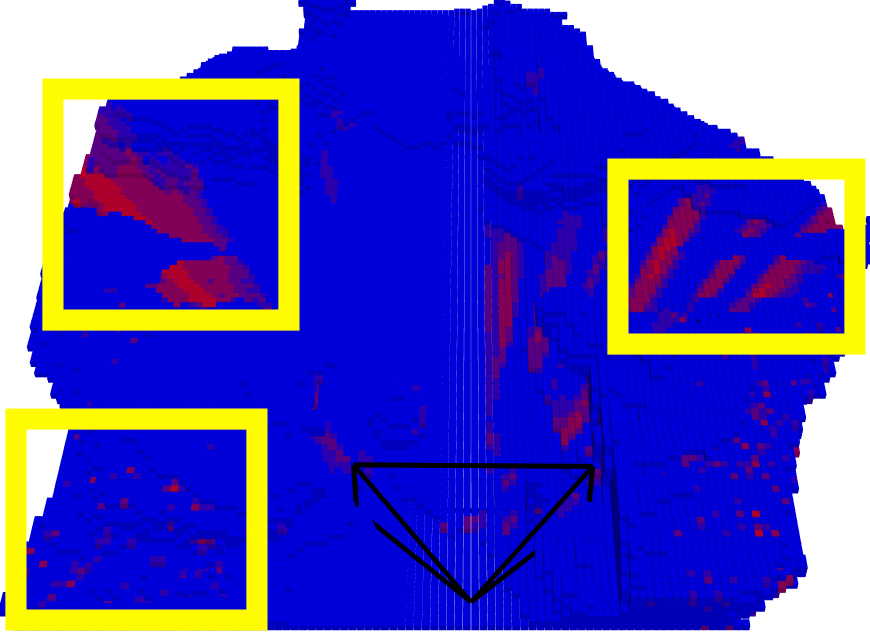} \end{minipage}
    \\ %\hline
    
    \begin{minipage}{.27\linewidth} \includegraphics[width=\linewidth]{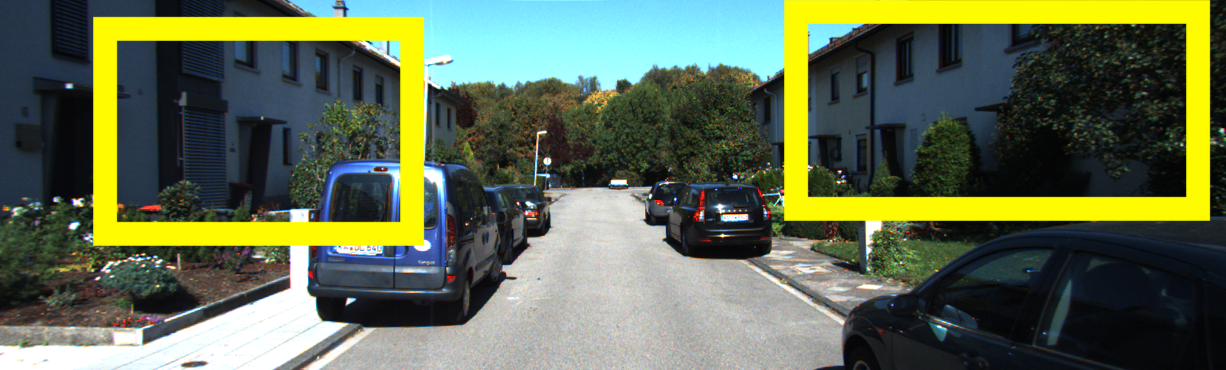} \end{minipage}
    &
    \begin{minipage}{.2\linewidth} \includegraphics[width=0.8\linewidth]{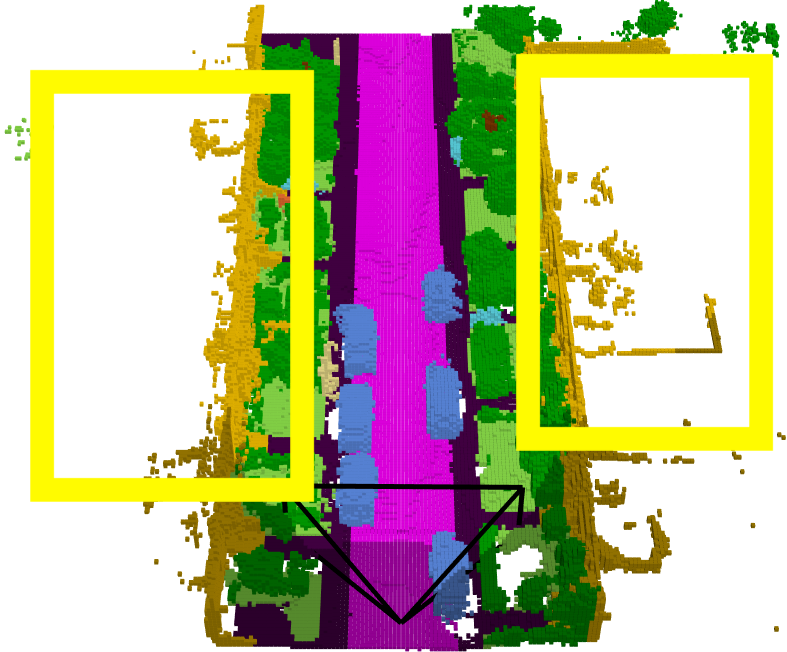} \end{minipage}
    &
    \begin{minipage}{.2\linewidth} \includegraphics[width=\linewidth]{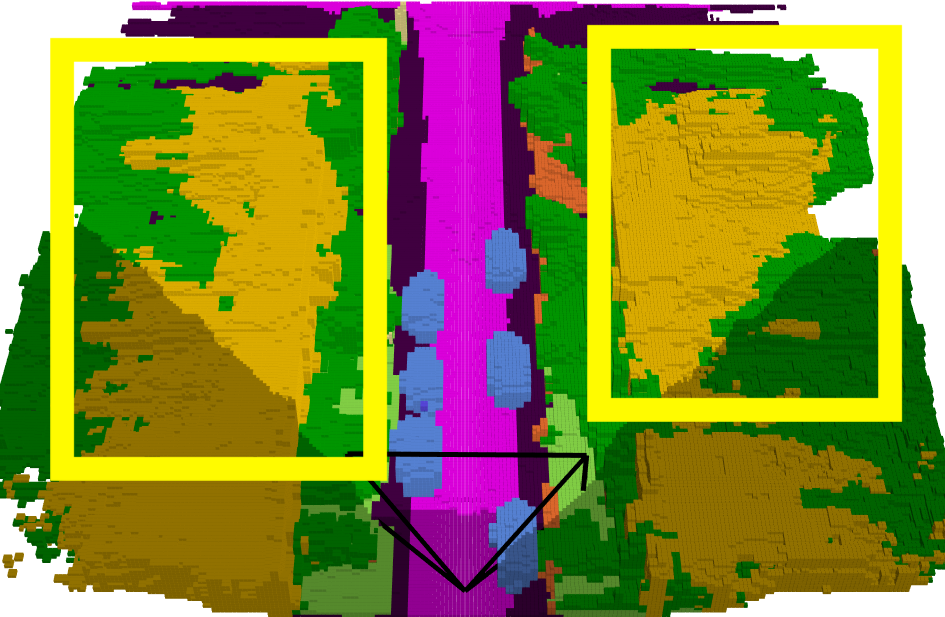} \end{minipage}
    &
    \begin{minipage}{.2\linewidth} \includegraphics[width=\linewidth]{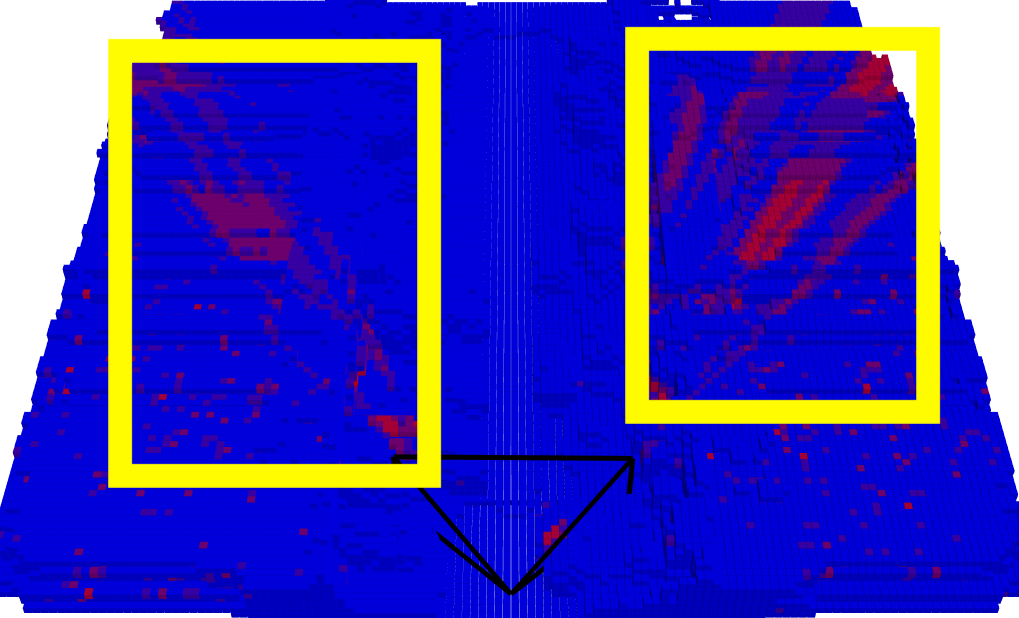} \end{minipage}
    \\ %\hline
  \end{tabular}
  \caption{%\footnotesize
  \textbf{Uncertainty Qualitative Results:} In column 4, blue voxels show low uncertainty areas, while red voxels, highlighted by yellow rectangles, indicate high uncertainty. These uncertain areas are typically outside the camera's FOV, or obscured by vegetation (row 1) or buildings (rows 2-3). }
  \label{fig:uncertainty_qualitative}
  % \vspace{-2em}
\end{figure*}

\subsection{Occupancy Prediction} \label{sec:stage1}
\begin{figure}[H]
    \centering
    \includegraphics[width=0.9\linewidth]{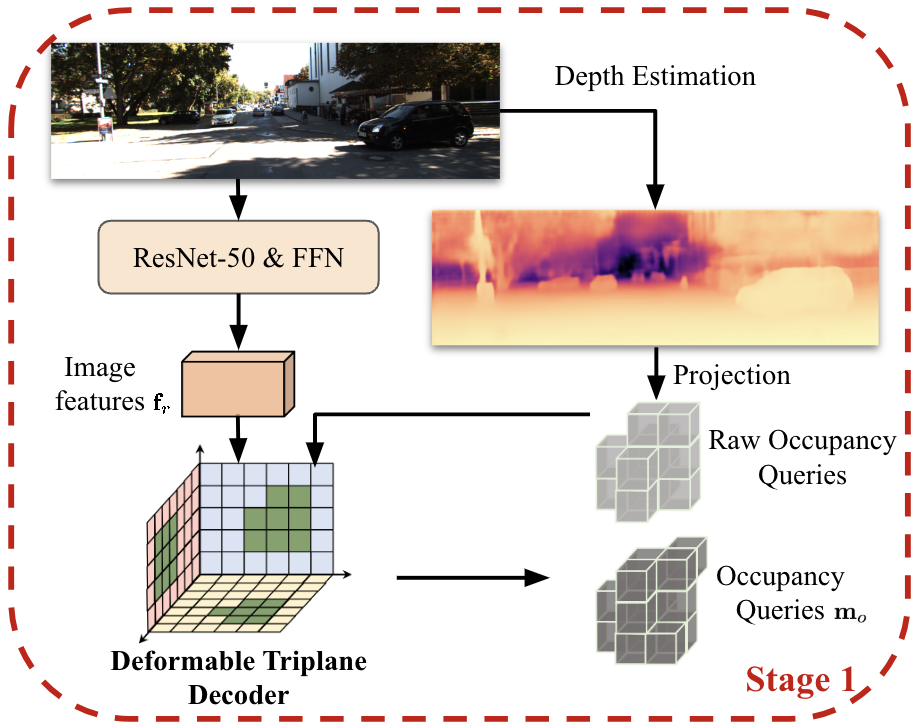}
    \caption{%\footnotesize
    \textbf{Stage 1}: We employ the novel triplane-based deformable attention method in this stage. The estimated depth pixels are projected to the 3D space as raw occupancy queries to the triplane-based deformable model. The output of stage~1 is an occupancy map, the occupied voxels in which are then converted to occupancy queries.}
    \label{fig:stage1}
    \vspace{-1em}
\end{figure}

In stage~1, the occupancy map $\m_o$ is estimated, which identifies occupied voxels in 3D space. As shown in Figure~\ref{fig:stage1}, we estimate depth from the RGB image, and then the depth pixels are projected into 3D space, voxelized into a raw occupancy map, and finally converted to raw occupancy queries through positional embedding. The initial map is often inaccurate due to raw depth estimation. Following a similar approach to stage~2, we use ResNet-50 and FPN to process image features, enhancing the occupancy map with the triplane-based deformable attention model described in Section~\ref{sec:tda}. In the triplane model, we use raw occupancy query as the query features and aggregate them into three orthogonal planes. The output of the triplane model is an occupancy map $\m_o = \l_o([\f_\text{cross}^l,\f_\text{cross}^h,\f_\text{cross}^d])$, where $\l_o$ is a linear layer. 

\subsection{Training Strategy}
The two stages are trained separately, using a cosine scheduler. Given that the ground truth occupancy map $\m_o^{gt}$ is derived from a sequence of LIDAR data, which may include occlusions, we mask out missing and unlabeled voxels during training. Stage~1 uses a cross-entropy loss $\cl_{ce}(\m_{o}^{gt}, \m_o)$ for occupancy prediction.

Stage~2 is trained in the formulation of CVAE, so we have the CVAE lower bound loss function:
\begin{align}
    \cl_{C} = -\cd_{KL} (p_\theta (\z|\m_o, \f_r) \parallel p (\z)) + \sum \log p(\m_s|\f_r,\z),
    \nonumber
\end{align}
where $p_\theta(\z|\m_o, \f_r)$ represents the approximate posterior distribution parameterized by the Voxel Encoder, and the KL divergence trains the distribution close to, $p(\z)$, the Gaussian distribution of $\z$. Note that the posterior distribution $p_\theta (\z|\m_o, \f_r)$ was used to sample $\hat{\f}_o$ during training as Equation \ref{eq:gaussian}. $p(\m_s|\f_r,\z)$ represents the conditional distribution of generating the semantic map $\m_s$ from the latent sample $\z$ and image feature $\f_r$. Besides KL divergence, the reconstruction loss includes class-weighted cross-entropy loss $\cl_{ce}$, and losses $\cl_{scal}^{sem}$ and $\cl_{scal}^{geo}$ from \cite{cao2022monoscene}.

\section{Experiments}
\textbf{Dataset and Evaluation Metrics:} In this work, various experiments were conducted to evaluate the proposed 3D uncertainty semantic prediction framework using the SemanticKITTI \cite{behley2019semantickitti} dataset with monocular cameras. SemanticKITTI provides the semantic ground truth data in the range of $[0 \ \text{m}, 51.2 \ \text{m}] \times [-25.6 \ \text{m}, 25.6 \ \text{m}]\times [-2 \ \text{m},4.4 \ \text{m}]$. The semantic occupancy map has the shape of $256\times256\times 32$ voxels, while the occupancy map for stage~1 has the shape $128 \times 128 \times 16$. Each voxel in the semantic occupancy map has 20 classes of semantic labels consisting of 19 semantic classes and 1 free class. Following the regular metric for SSC tasks, we utilize the Intersection over Union (IoU) to evaluate the occupancy prediction and the scene completion quality in stages one and two. For semantic evaluation, we use the mean IoU (mIoU) of the 19 semantic classes to evaluate the performance of the semantic prediction. Furthermore, we enumerate all 19 classes to evaluate the accuracy of the methods in each semantic class estimation. The training and evaluation are done by one GPU V100.

% \textbf{Comparisons:}
% We compare with OccFormer~\cite{zhang2023occformer}, VoxFormer~\cite{li2023voxformer}, 

\subsection{Comparisons}
Table~\ref{tab:ssc} compares our approach, ET-Former, with state-of-the-art methods. Our model achieves the highest IoU and mIoU, surpassing other methods by improving the SOTA scores of IoU by $15.16\%$ (from $44.71$ to $51.49$) and mIoU by $8.38\%$ (from $15.04$ to $16.30$). The triplane-based feature representation proves particularly effective for commonly seen, large semantic objects. We achieve the best semantic accuracy across multiple categories, such as building, car, other-vehicle, vegetation, trunk, terrain, and pole.

We evaluated the  training memory usage of ET-Former against other approaches (Table~\ref{tab:memory}). ET-Former reduces GPU memory usage by 3.7~GB $(25\%)$ compared to VoxFormer, and by 11.1 GB $(50.5\%)$ compared to Symphonies~\cite{jiang2024symphonize}, while significantly outperforming them in IoU and mIoU. These results highlight the efficiency of the proposed method.
\begin{table}[H]
    \centering
    \begin{tabular}{l|l|l|l}
      %  Methods  & Model Size (Mb) & Inference GPU Usage (Mb) \\ \hline
      % VoxFormer~\cite{li2023voxformer}  &233.1 & 1867\\ \hline
      %  MonoScene~\cite{cao2022monoscene} & 600.1 & 2013\\ \hline
      %  ET-Former  & 319.5 & 1133
        Methods & IoU & mIoU & Memory  \\ \hline
      VoxFormer~\cite{li2023voxformer} & 42.95 & 12.20 &14.6 Gb \\ 
       MonoScene~\cite{cao2022monoscene} & 34.16 & 11.08 & 18.9 Gb \\ 
       Symphonies ~\cite{jiang2024symphonize} & 42.19 & 15.04 & 22.0 Gb  \\ 
       TPVFormer~\cite{huang2023tri} &34.25 &11.26 & 23.0 Gb \\ \hline
       ET-Former  & \textbf{51.49} & \textbf{16.30} & \textbf{10.9} Gb 
    \end{tabular}
    
    \caption{%\footnotesize
    \textbf{Training Memory Usage On SemanticKITTI}: Our method reduces GPU memory usage by 3.7 GB $(25\%)$ compared to VoxFormer~\cite{li2023voxformer}, and by 11.1 GB $(50.5\%)$ compared to Symphonies~\cite{jiang2024symphonize}, while outperforming them in IoU and mIoU.}
    \label{tab:memory}
    \vspace{-1.5em}
\end{table}

\begin{figure}[!ht]
  \centering
  \begin{tabular}{ c c c }
  % \hline
   \begin{minipage}{.01\linewidth}
   \rotatebox{90}{\textbf{RGB} }\end{minipage}
   & 
   \begin{minipage}{.44\linewidth} \includegraphics[width=\linewidth]{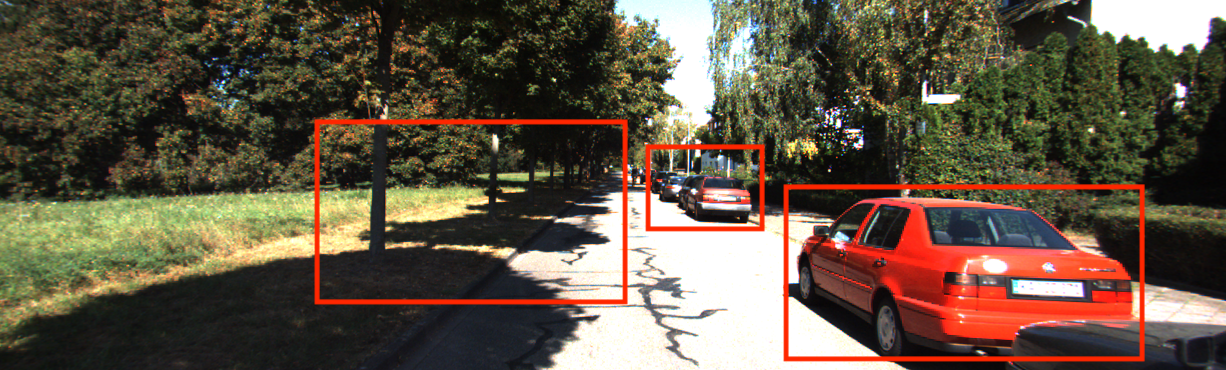} \end{minipage}
   & 
   \begin{minipage}{.44\linewidth} \includegraphics[width=\linewidth]{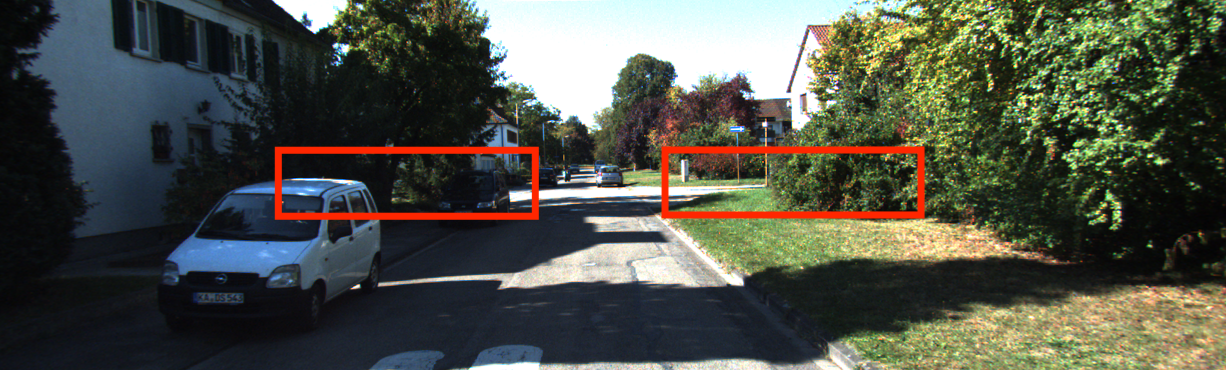} \end{minipage}
   \\ 
   % \hline
   \begin{minipage}{.01\linewidth}
   \rotatebox{90}{\textbf{VoxFormer~\cite{li2023voxformer}} }\end{minipage}
    & 
   \begin{minipage}{.44\linewidth} \includegraphics[width=\linewidth]{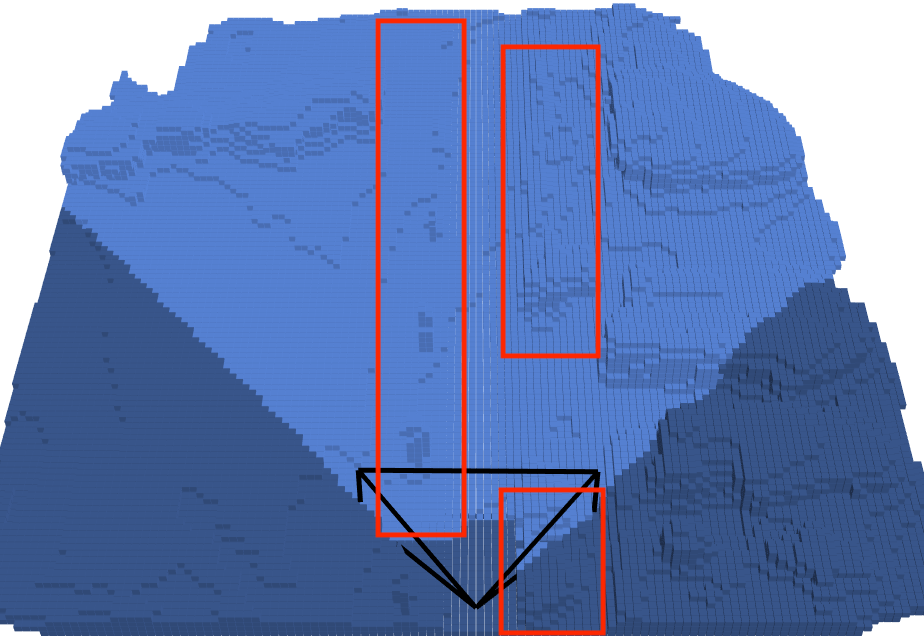} \end{minipage}
   &
   \begin{minipage}{.44\linewidth} \includegraphics[width=\linewidth]{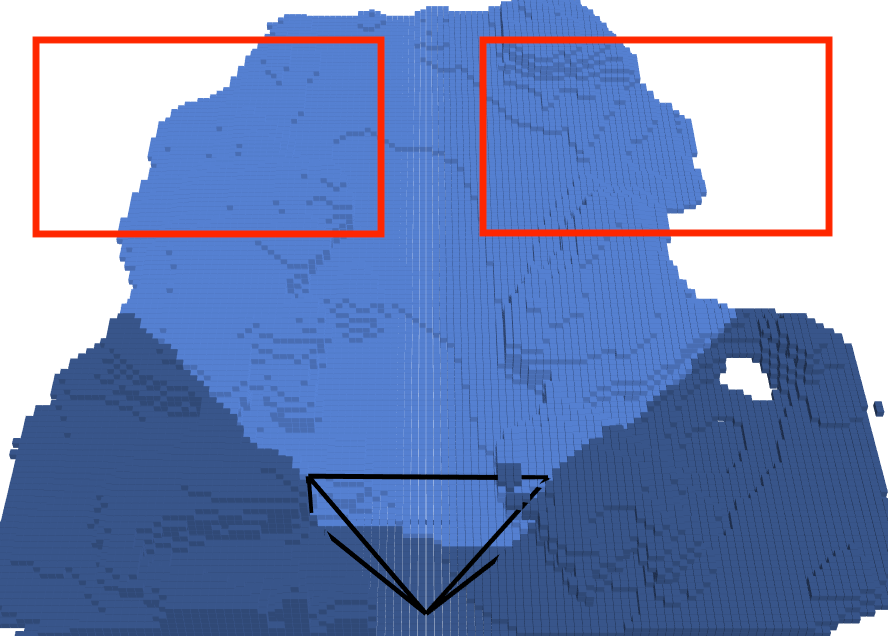} \end{minipage}
   \\ 
   % \hline
   \begin{minipage}{.01\linewidth}
   \rotatebox{90}{\textbf{ET-Former} }\end{minipage} 
   &
   \begin{minipage}{.44\linewidth} \includegraphics[width=\linewidth]{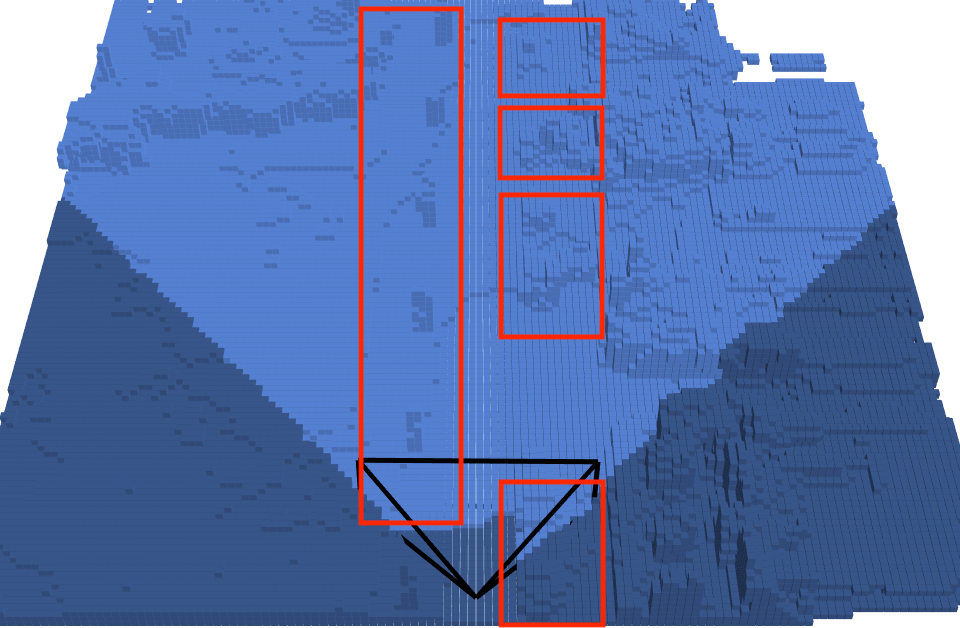} \end{minipage}
   &
   \begin{minipage}{.44\linewidth} \includegraphics[width=\linewidth]{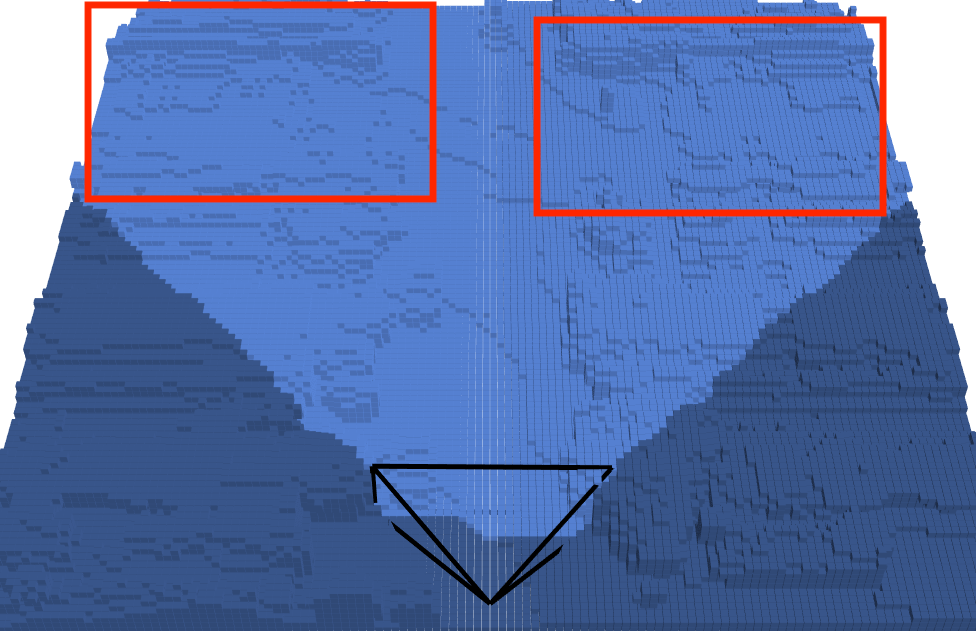} \end{minipage}
    \\ 
    % \hline
  \end{tabular}
  \caption{%\footnotesize
  \textbf{Stage 1 Restuls:} The major differences between the stage~1 results of ET-Former and VoxFormer are marked by red rectangles. In the first column, ET-Former demonstrates superior performance in detecting trees and cars, providing more detailed and accurate representations. The second column illustrates ET-Former's ability to generate a more comprehensive map, including the detection of distant roads that VoxFormer has missed.}
  \label{fig:stage1_results}
  \vspace{-1.5em}
\end{figure}

To evaluate our stage~1 occupancy prediction model proposed in Section~\ref{sec:stage1}, we conducted a comparative analysis with VoxFormer~\cite{li2023voxformer}. To ensure a fair comparison, we utilize the same depth estimation method, MobileStereoNet~\cite{shamsafar2022mobilestereonet}, as employed by the stage~1 of VoxFormer. Subsequently, as depicted in Figure~\ref{fig:stage1}, we apply our triplane-based method to generate the occupancy map and queries. The quality of the occupancy map predictions is compared between our stage~1 approach and VoxFormer's, as shown in Table~\ref{tab:stage1}. Our approach shows substantial improvements, with a $14.2\%$ increase in IoU and a $16.3\%$ increase in Recall, resulting in a more accurate occupancy map and more efficient occupancy queries. Figure~\ref{fig:stage1_results} shows the qualitative comparison between our stage~1 approach and VoxFormer's. We observe that our approach, ET-Former, can estimate details more accurately and generate more complete occupancy map than VoxFormer.  The improvement in stage~1 ultimately contributes to the performance improvement of the subsequent semantic predictions, as shown in Table~\ref{tab:ablation}. % \TODO{we may want to explain why the IoU is larger than Stage2: the first stage only uses 128x128x16 dimension, but stage 2 uses 256x256x32. the total size is larger and therefore harder.}

\begin{table}[H]
    \centering
    \begin{tabular}{l|l|l}
      Methods       & IoU   & Recall \\ \hline
      VoxFormer's stage~1     & 52.03 & 61.52 \\ \hline
      Our stage~1 & 59.41 & 71.56 \\ 
    \end{tabular}
    \caption{\textbf{Results of Stage~1 On SemanticKITTI \texttt{test} \cite{behley2019semantickitti}}: Compared with the VoxFormer's stage~1 model, our stage~1 model achieves $14.2\%$ and $16.3\%$ improvements in IoU and Recall values, respectively.}
    \label{tab:stage1}
    \vspace{-2em}
\end{table}

    % \begin{minipage}{.2\linewidth} \includegraphics[width=\linewidth]{figs/uncertainty/925_uncertainty.png} \end{minipage}

\subsection{Ablation Study}
Table \ref{tab:ablation} presents an ablation analysis of the architectural components, using VoxFormer~\cite{li2023voxformer} as the baseline. Replacing VoxFormer's stage~2 with our deterministic stage~2 model (where query features $\f_o$ are directly fed into the deformable triplane decoder instead of converting to Gaussian samples $\hat{\f}_o$ as in Equation \ref{eq:gaussian}) improves IoU by 0.36 and mIoU by 0.28 compared to the baseline. Implementing our CVAE stochastic stage~2 further improves performance by 0.63 IoU and 0.57 mIoU. Finally, combining our stage~1 model with the CVAE stochastic  stage~2 significantly boosts performance by 7.57 IoU and 3.25 mIoU.

% To evaluate the impact of the CVAE in semantic estimation, where stochastic latent Gaussian distributions is capable of constraining the latent embeddings, we compare the deterministic model with the CVAE stochastic training method in Table \ref{tab:ablation}. The CVAE stochastic approach enhances the accuracy of both semantic and occupancy estimations, leading to at least a 0.63 improvement in IoU and a 0.57 improvement in mIoU. 

\begin{table}[!h]
    \centering
    \begin{tabular}{l|l|l}
        Methods & IoU & mIoU \\ \hline
        Baseline (VoxFormer~\cite{li2023voxformer}) & 42.95 & 12.20 \\ \hline
        % OccFormer~\cite{zhang2023occformer} & 34.53 & 12.32 \\ \hline
        + Our deterministic stage 2 & 43.29 \textcolor{gray}{(+0.36)} & 12.48 \textcolor{gray}{(+0.28)}\\
        + Our stochastic stage 2 & 43.92 \textcolor{gray}{(+0.63)} & 13.05 \textcolor{gray}{(+0.57)} \\
        + Our stage 1 + stochastic stage 2 & \textbf{51.49} \textcolor{gray}{(+7.57)} & \textbf{16.30} \textcolor{gray}{(+3.25)}
    \end{tabular}
    \caption{%\footnotesize
    \textbf{Ablation Study On SemanticKITTI \texttt{test} \cite{behley2019semantickitti}:} our key components (deformable triplane decoder, CVAE formulation, and proposed stage~1 model) each contribute to significant improvements in semantic estimation accuracy.}
    \label{tab:ablation}
        \vspace{-1.5em}
\end{table}

\subsection{Visualizations}
\textbf{Qualitative Results:} Figure~\ref{fig:semantic_qualitative} qualitatively compares our approach with VoxFormer~\cite{li2023voxformer} and MonoScene~\cite{cao2022monoscene}. Blue rectangles highlight areas present in the Ground Truth but missed by VoxFormer and MonoScene. Red rectangles indicate regions where VoxFormer and MonoScene made incorrect semantic estimations. VoxFormer shows considerable noise in its predictions, especially outside the camera's FOV, and struggles with occluded areas (e.g., the driving road in the third row). MonoScene has difficulty estimating semantics in occluded areas and completing the map. In contrast, our ET-Former effectively handles occluded and out-of-FOV areas.

\textbf{Uncertainty Map visualization:} In Figure~\ref{fig:uncertainty_qualitative}, we evaluate the uncertainties in the estimated semantic occupancy map. The yellow rectangles highlight areas with high uncertainty, which mainly arise from occlusions and regions outside the camera's FOV. For example, in the first and third rows, areas behind vegetation and buildings exhibit significant uncertainties, respectively, with the uncertainty extending outward from these objects along the camera's line of sight. In the second row, while the horizontal road is blocked by inaccurately estimated terrain, our method assigns high uncertainty values to these problematic areas. This uncertainty quantification helps identify regions where the occupancy estimation may be less reliable, potentially aiding in formulation of risk-aware navigation strategies.

\section{Conclusion, Limitations, and Future Work}
% \textbf{Conclusion:} 
ET-Former offers an efficient and robust solution for 3D semantic scene completion using a single monocular camera. It successfully achieves state-of-the-art performance with much less memory usage, highlighting its potential for real-world applications in autonomous navigation systems. Our future work could focus on designing a more memory-efficient yet effective encoder. Additionally, we aim to design a navigation strategy that makes safe but also permissible decisions leveraging both semantic and uncertainty maps generated from the proposed algorithm.

\bibliographystyle{IEEEtran}
\bibliography{ref}

% \clearpage
\newpage

\end{document}